\documentclass[10pt, sigconf,table]{acmart}
\renewcommand\footnotetextcopyrightpermission[1]{} 
\usepackage{booktabs} 
\usepackage{times}
\usepackage{algorithm}
\usepackage[noend]{algorithmic}
\usepackage{enumitem}
\usepackage{mathtools}
\usepackage{listings}
\usepackage{tabularx}
\usepackage{adjustbox}
\usepackage{array}
\usepackage{float}
\usepackage{graphicx}
\usepackage{epstopdf}
\usepackage{color}
\usepackage{url}
\usepackage{xspace}
\usepackage{balance}
\usepackage{amsmath,amssymb,graphicx}
\usepackage[short]{optidef}
\usepackage{microtype}
\usepackage{sepfootnotes}
\usepackage[margin=5pt,font=small,labelfont={bf,rm}]{caption}

\renewcommand{\paragraph}[1]{\smallskip\noindent\textbf{#1}}

\makeatletter
\g@addto@macro{\UrlBreaks}{\UrlOrds}
\makeatother

\newif\ifgmlegal
\gmlegalfalse

\newcommand{\sys}{Grab\xspace}

\makeatletter
\def\maxwidth{\ifdim\Gin@nat@width>\linewidth\linewidth\else\Gin@nat@width\fi}
\def\maxheight{\ifdim\Gin@nat@height>\textheight\textheight\else\Gin@nat@height\fi}
\makeatother
\setkeys{Gin}{width=\maxwidth,height=\maxheight,keepaspectratio}
\renewcommand{\footnotesize}{\scriptsize}

\makeatletter
\let\origsection\section
\let\origsubsection\subsection

\renewcommand\section{\@ifstar{\starsection}{\nostarsection}}
\renewcommand\subsection{\@ifstar{\starsubsection}{\nostarsubsection}}

\newcommand\sectionprelude{\vspace{-0.75ex}}
\newcommand\sectionpostlude{\vspace{-0.5ex}}
\newcommand\subsectionprelude{\vspace{-0.5ex}}
\newcommand\subsectionpostlude{\vspace{-0.5ex}}

\newcommand\nostarsection[1]{\sectionprelude\origsection{#1}\sectionpostlude}
\newcommand\starsection[1]{\sectionprelude\origsection*{#1}\sectionpostlude}

\newcommand\nostarsubsection[1]{\subsectionprelude\origsubsection{#1}\subsectionpostlude}
\newcommand\starsubsection[1]{\subsectionprelude\origsubsection*{#1}\subsectionpostlude}

\makeatother

\newcommand\paraspace{\vspace*{0.5ex}}

\newcommand{\parab}[1]{\paraspace\noindent\textbf{#1}.}
\newcommand{\parae}[1]{\paraspace\textbf{\emph{#1}.}}

\definecolor{brown}{cmyk}{0,0.81,1,0.60}
\definecolor{magenta}{rgb}{0.4,0.7,0}
\definecolor{gray}{rgb}{0.5,0.5,0.5}
\definecolor{red}{rgb}{1,0,0}
\definecolor{green}{rgb}{0.5,0,0.5}
\definecolor{blue}{rgb}{0,0,1}

\newcommand{\etc}{\emph{etc.}\xspace}
\newcommand{\ie}{\emph{i.e.,}\xspace}
\newcommand{\eg}{\emph{e.g.,}\xspace}

\newcommand{\secref}[1]{\S\ref{#1}}

\newcommand{\figref}[1]{Figure~\ref{#1}}
\newcommand{\tblref}[1]{Table~\ref{#1}}

\def\setof#1{\left\{{\let\st\colon #1 }\right\}}

    \def\00{\mathbf{0}}

\newcommand{\eat}[1]{}

\newcolumntype{L}[1]{>{\raggedright\let\newline\\\arraybackslash\hspace{0pt}}m{#1}}
\newcolumntype{C}[1]{>{\centering\let\newline\\\arraybackslash\hspace{0pt}}m{#1}}
\newcolumntype{R}[1]{>{\raggedleft\let\newline\\\arraybackslash\hspace{0pt}}m{#1}}

\begin{document}

\title{\sys: Fast and Accurate Sensor Processing for Cashier-Free Shopping}

\author{Xiaochen Liu}
\affiliation{\institution{University of Southern California}}
\email{liu851@usc.edu}

\author{Yurong Jiang}
\affiliation{\institution{LinkedIn*}}
\email{jiangyurong609@gmail.com}

\author{Kyu-Han Kim}
\affiliation{\institution{Hewlett-Packard Labs}}
\email{kyu-han.kim@hpe.com}

\author{Ramesh Govindan}
\affiliation{\institution{University of Southern California}}
\email{ramesh@usc.edu}


\begin{abstract}
Cashier-free shopping systems like Amazon Go improve shopping experience, but can require significant store re-design. In this paper, we propose \sys, a practical system that leverages existing infrastructure and devices to enable cashier-free shopping. \sys needs to accurately identify and track customers, and associate each shopper with items he or she retrieves from shelves. To do this, it uses a keypoint-based pose tracker as a building block for identification and tracking, develops robust feature-based face trackers, and algorithms for associating and tracking arm movements. It also uses a probabilistic framework to fuse readings from camera, weight and RFID sensors in order to accurately assess which shopper picks up which item. In experiments from a pilot deployment in a retail store, \sys can achieve over 90\% precision and recall even when 40\% of shopping actions are designed to confuse the system.  Moreover, \sys has optimizations that help reduce investment in computing infrastructure  four-fold.

\end{abstract}

\thanks{* The work was done at Hewlett-Packard Labs}

\maketitle


\section{Introduction}
\label{sec:intro}

While electronic commerce continues to make great strides, in-store purchases are likely to continue to be important in the coming years: 91$\%$ of purchases are still made in physical stores \cite{forrester, phy_shopping} and 82$\%$ of millennials prefer to shop in these stores \cite{millennial}. However, a significant pain point for in-store shopping is the checkout queue: customer satisfaction drops significantly when queuing delays exceed more than four minutes \cite{checkout_time}. To address this, retailers have deployed self-checkout systems (which can increase instances of shoplifting \cite{selfcheckout_time, selfcheckout_theft1, selfcheckout_theft2}), and expensive vending machines.

The most recent innovation is \textit{cashier-free shopping}, in which a networked sensing system automatically (a) \textit{identifies} a customer who enters the store, (b) \textit{tracks} the customer through the store, (c) and \textit{recognizes} what they purchase. Customers are then billed automatically for their purchases, and do not need to interact with a human cashier or a vending machine, or scan items by themselves. Over the past year, several large online retailers like Amazon and Alibaba~\cite{amazon, taobao} have piloted a few stores with this technology, and cashier-free stores are expected to take off in the coming years~\cite{cashierfreetrend,retaileradopt}. Besides addressing queue wait times, cashier-free shopping is expected to reduce instances of theft, and provide retailers with rich behavioral analytics.

\sepfootnotecontent{grab}{A shopper only needs to \textit{grab} items and go.}

Not much is publicly known about the technology behind cashier-free shopping, other than that stores need to be completely redesigned~\cite{amazon, taobao, bingobox} which can require significant capital investment (\secref{sec:motivation}). In this paper, we ask: Is cashier-free shopping viable without having to completely redesign stores? To this end, we observe that many stores already have, or will soon have, the hardware necessary to design a cashier-free shopping system: cameras deployed for in-store security, sensor-rich smart shelves~\cite{smart_shelves} that are being deployed by large retailers~\cite{smallbig} to simplify asset tracking, and RFID tags being deployed on expensive items to reduce theft. Our paper explores the design and implementation of a practical cashier-free shopping system called \sys\sepfootnote{grab} using this infrastructure, and quantifies its performance.

\sys needs to accurately identify and track customers, and associate each shopper with items he or she retrieves from shelves. It must be robust to visual occlusions resulting from multiple concurrent shoppers, and to concurrent item retrieval from shelves where different types of items might look similar, or weigh the same. It must also be robust to fraud, specifically to attempts by shoppers to confound identification, tracking, or association. Finally, it must be cost-effective and have good performance in order to achieve acceptable accuracy: specifically, we show that, for vision-based tasks, slower than 10 frames/sec processing can reduce accuracy significantly (\secref{sec:eval}). 

\parab{Contributions}
An obvious way to architect \sys is to use deep neural networks (DNNs) for each individual task in cashier-free shopping, such as identification, pose tracking, gesture tracking, and action recognition. However, these DNNs are still relatively slow and many of them cannot process frames at faster than 5-8 fps. Moreover, even if they have high individual accuracy, their effective accuracy would be much lower if they were cascaded together.

\sys's architecture is based on the observation that, for cashier-free shopping, we can use a single vision capability (body pose detection) as a building block to perform \textit{all} of these tasks. A recently developed DNN library, OpenPose~\cite{openpose} accurately estimates body "skeletons" in a video at high frame rates. 

\sys's first contribution is to develop a suite of lightweight identification and tracking algorithms built around these skeletons (\secref{sec:identity}). \sys uses the skeletons to accurately determine the bounding boxes of faces to enable feature-based face detection. It uses skeletal matching, augmented with color matching, to accurately track shoppers even when their faces might not be visible, or even when the entire body might not be visible. It augments OpenPose's elbow-wrist association algorithm to improve the accuracy of tracking hand movements which are essential to determining when a shopper may pickup up items from a shelf.

\sys's second contribution is to develop fast sensor fusion algorithms to associate a shopper's hand with the item that the shopper picks up (\secref{sec:behavior}). For this, \sys uses a  probabilistic assignment framework: from cameras, weight sensors and RFID receivers, it determines the likelihood that a given shopper picked up a given item. When multiple concurrent such actions occur, it uses an optimization framework to associate hands with items.

\sys's third contribution is to improve the cost-effectiveness of the overall system by multiplexing multiple cameras on a single GPU (\secref{sec:scale}). It achieves this by avoiding running OpenPose on every frame, and instead using a lightweight feature tracker to track the joints of the skeleton between successive frames.

Using data from a pilot deployment in a retail store, we show (\secref{sec:eval}) that \sys has  93\% precision and 91\% recall even when nearly 40\% of shopper actions were adversarial. \sys needs to process video data at 10 fps or faster, below which accuracy drops significantly: a DNN-only design cannot achieve this capability (\secref{sec:efficiency_eval}). \sys needs all three sensing modalities, and all of its optimizations: removing an optimization, or a sensor, can drop precision and recall by 10\% or more. Finally, \sys's design enables it to multiplex up to 4 cameras per GPU with negligible loss of precision.

\section{Approach and Challenges}
\label{sec:motivation}

\begin{figure}
\centering\includegraphics[width=0.85\columnwidth]{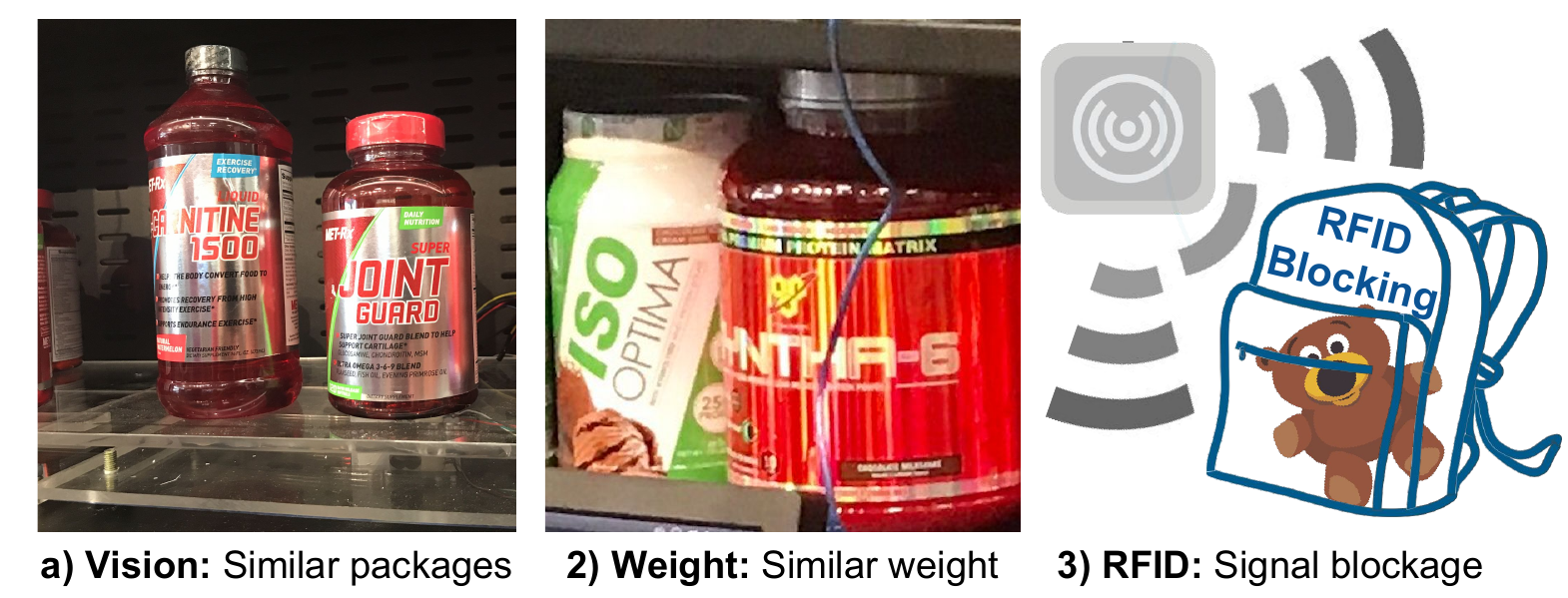}
\caption{\emph{Real-world challenges in identifying items. a) Different items with similar packages; b) Different items with similar weight; c) Occluded RFID tags that would be hard to read at a checkout gate.}}
\label{fig:motivation}
\end{figure}




\parab{Cashier-free shopping systems} A cashier-free shopping system \textit{automatically} determines, for every customer in a shop, what \textit{items} the customer has picked from the shelves, and directly bills each customer for those items. Cashier-free shopping is achieved using a networked system containing several sensors that together perform three distinct functions: identifying each customer, tracking each customer through the shop, and identifying every \textit{item pickup} or \textit{item dropoff} on a shelf to accurately determine which items the customer leaves the store with.

These systems have several requirements. First, they must be \textit{non-intrusive} in the sense that they must not require customers to wear or carry sensors or any form of electronic identification, since these can detract from the shopping experience. Second, they must be \textit{robust to real-world conditions} (\figref{fig:motivation}), in being able to  distinguish between items that are visually similar or have other similarities (such as weight), as well as to be robust to occlusion. Third, they must be \textit{robust to fraud}: specifically, they must be robust to attempts by shoppers to circumvent or tamper with sensors used to identify customers, items and the association between customers and items. Finally, they must be \textit{cost-effective}: they should leverage existing in-store infrastructure to the extent possible, while also being computationally efficient in order to minimize computing infrastructure investments.

\parab{Today's cashier-free shopping systems} Despite widespread reports of cashier-free shopping deployments~\cite{amazon, taobao, bingobox, std_cog}, not much is known about the details of their design, but they appear to fall into three broad categories.

\parae{Vision-Only}
This class of systems, exemplified by \cite{std_cog_forbes,std_cog}, identifies customers and items, and tracks customers, only using cameras. It trains a deep learning model to recognize customers and objects, and uses this to bill them. However, such a system can fail to distinguish between items that look similar (\figref{fig:motivation}(a)) especially when these items are small in the image (occupy a few pixels), or items that are occluded by other objects (\figref{fig:motivation}(c)) or by the customer.

\parae{Vision and Weight}
Amazon Go \cite{amazon} uses both cameras and weight sensors on shelves, where the weight sensor can be used to identify when an item is removed from a shelf even if it is occluded from a camera. One challenge such a system faces is the ability to discriminate between items of similar weight (\figref{fig:motivation}(b)). Moreover, their design requires a significant redesign of the store: user check-in gates, an array of cameras on the ceiling and the shelf, and additional sensors at the exit~\cite{go-nyt,go-geek}. Finally, Amazon Go also reportedly encounters issues when shoppers put back items randomly~\cite{amzn_prob}.

\parae{Vision and RFID}
The third class of approaches, used by Taobao Cafe~\cite{taobao} and Bingo Box~\cite{bingobox}, does not track shoppers within the store, but uses vision to identify customers and RFID scanners at a checkout gate that reads all the items being carried by the customer. Each object needs to be attached with an RFID tag, and users have to queue at the checkout gate. This approach has drawbacks as well: RFID tags can be expensive relative to the price of some items~\cite{rfid_tag}, and RFID readers are known to have trouble when scanning tags that are stacked, blocked, or attached to conductors \cite{nikitin2006performance, lu2009performance, nekoogar2011ultra}. 


\parab{Approach and Challenges} While all of these approaches are non-intrusive, it is less clear how well they satisfy other requirements: robustness to real-world conditions and to fraud, and cost-effectiveness. In this paper, with a goal towards understanding how well these requirements can be met in practice, we explore the design,  implementation, and evaluation of a cashier-free shopping system called \sys, which combines the three technologies described above (vision, weight scales, and RFID). At a high-level, \sys combines advances in machine vision, with lightweight sensor fusion algorithms to achieve its goals. It must surmount four distinct challenges: (a) how to identify customers in a lightweight yet robust manner; (b) how to track customers through a store even when the customer is occluded by others or the customer's face is not visible in a camera; (c) how to determine when a customer has picked up an item, and which item the customer has picked up, and to make this determination robust to concurrent item retrievals, customers putting back items, and customers attempting to game the system in various ways; (d) how to meet these challenges in a way that minimizes investments in computing infrastructure.

\section{\sys Design}
\label{sec:design}

\begin{figure}
\centering\includegraphics[width=0.85\columnwidth]{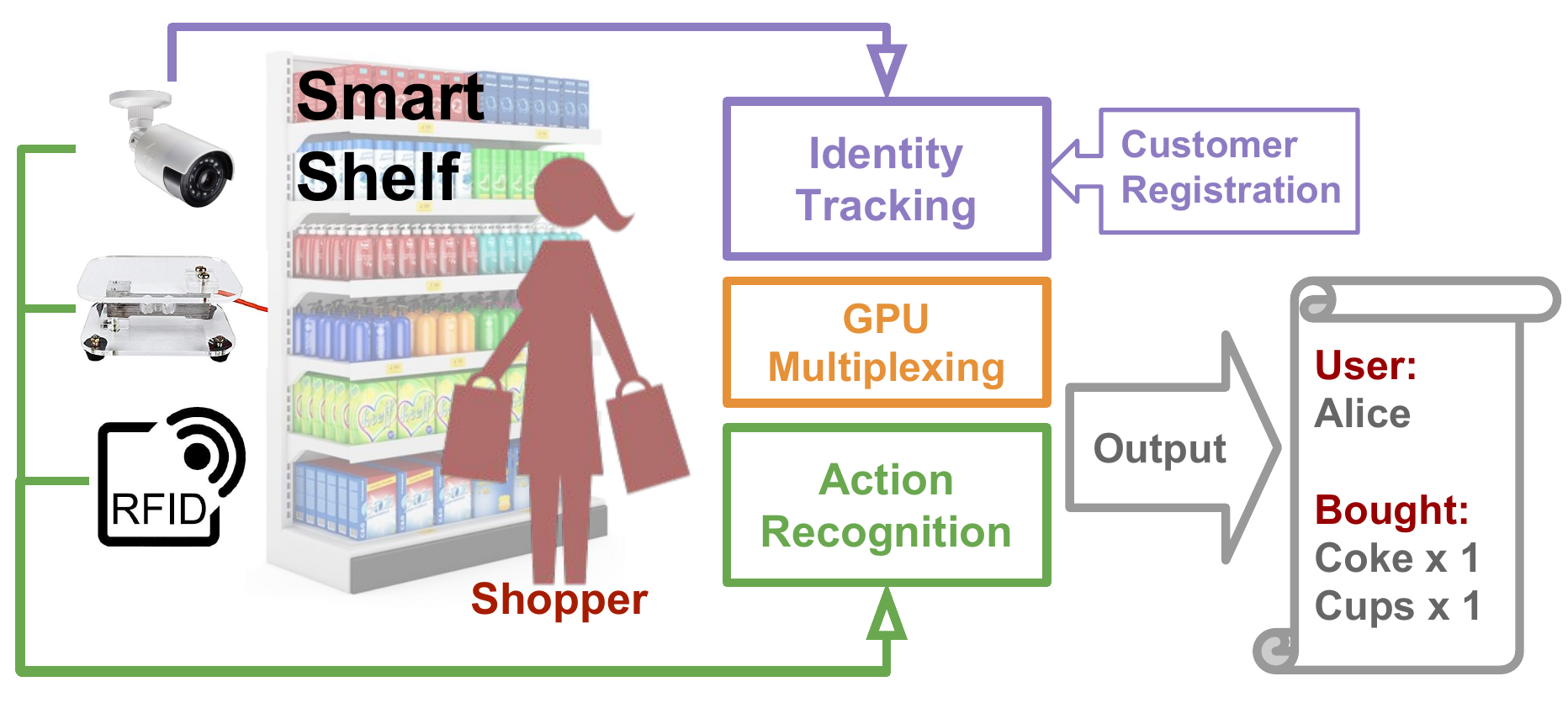}
\caption{\emph{\sys is a system for cashier-free shopping and has four components: registration, identity tracking, action recognition, and GPU multiplexing.}}
\label{fig:sys_design}
\end{figure}

\sys addresses these challenges by building upon a vision-based keypoint-based pose tracker DNN for identification and tracking, together with a probabilistic sensor fusion algorithm for recognizing item pickup actions. These ensure a completely non-intrusive design where shoppers are not required to scan item codes or pass through checkout gates while shopping. \sys consists of four major components (\figref{fig:sys_design}).

\textit{Identity tracking} recognizes shoppers' identities and tracks their movements within the store. It includes efficient and accurate face and body pose detection and tracking, adapted to work well in occluded environments, and to deal with corner cases in body pose estimation that can increase error in item pickup detection (\secref{sec:identity}).

\textit{Action recognition} uses a probabilistic algorithm to fuse vision, weight and RFID inputs to determine item pickup or dropoff actions by a customer (\secref{sec:behavior}). This algorithm is designed to be robust to real-world conditions and to theft. When multiple users pickup the same type of item simultaneously, the algorithm must determine which customer takes how many items. It must be robust to: customers concealing items or to attempts to tamper with the sensors (\eg replacing an expensive item with an identically weighted item). 

\textit{GPU multiplexing} enables processing multiple cameras on a single GPU (\secref{sec:scale}). DNN-based video processing usually requires a dedicated GPU for each video stream for reasonable performance. Retail stores need tens of cameras, and \sys contains performance optimizations that permit it to multiplex the processing of multiple streams on a single GPU, thereby reducing cost.

\sys also has a fourth, offline component, \textit{registration}. Customers must register \textit{once} online before their first store visit. Registration involves taking a video of the customer to enable matching the customer subsequently (\secref{sec:identity}), in addition to obtaining information for billing purposes. If the identity tracking component detects a customer who has not registered, she may be asked to register before buying items from the store. 

\subsection{Identity tracking}
\label{sec:identity}

\begin{figure}
\centering\includegraphics[width=0.85\columnwidth]{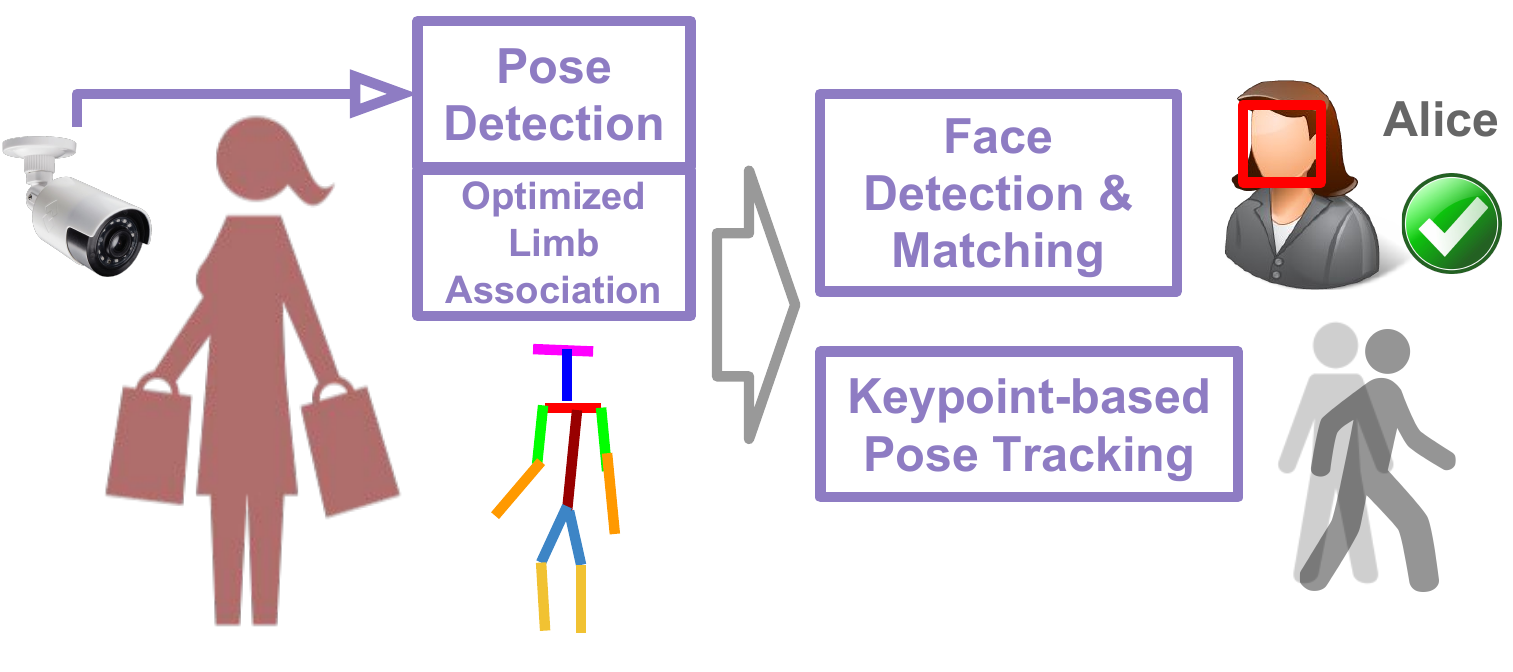}
\caption{\emph{\sys's identity tracking module uses a feature-based face detector and uses key-point base pose tracking.}}
\label{fig:identity_tracking}
\end{figure}

Identity tracking consists of two related sub-components (\figref{fig:identity_tracking}). Shopper identification determines \textit{who} the shopper is among registered users. A related sub-component, shopper tracking, determines (a) \textit{where} the shopper is in the store at each instant of time, and (b) \textit{what} the shopper is doing at each instant. 

\parab{Requirements and Challenges}
In designing \sys, we require first that customer registration be fast, even though it is performed only once: ideally, a customer should be able to register and immediately commence shopping. Identity tracking requires not just identifying the customer, but also detecting each person's \textit{pose}, such as hand position and head position. These tasks have been individually studied extensively in the computer vision literature. More recently, with advances in deep learning, researchers in computer vision have developed different kind of DNNs for people detection~\cite{ren2015faster, liu2016ssd, yolo}, face detection~\cite{zhang2016joint, dlib} and hand gesture recognition~\cite{tang2015real, chen2016deep}.

Each of these detectors performs reasonably well: \eg people detectors can process 35 frames per second (fps), face detectors can process 30 fps, and hand recognizers can run at 12 fps. However, \sys requires \textit{all of these components}. Dedicating a GPU for each component is expensive: recall that a store may have several cameras (\sys proposes to re-purpose surveillance cameras for visual recognition tasks, \secref{sec:intro}), and using one GPU per detection task per camera is undesirable (\secref{sec:motivation}) as it would require significant investment in computing infrastructure. 

The other option is to run these on a single GPU per camera, but this would result in lower frame rates. Lower frame rates can miss shopper actions: as \secref{sec:efficiency_eval} shows, at frame rates lower than 10 fps, \sys's precision and recall can drop dramatically. This highlights a key challenge we face in this paper: \textit{designing fast end-to-end identity tracking} algorithms that do not compromise accuracy.

\begin{figure}
\centering\includegraphics[width=0.70\columnwidth]{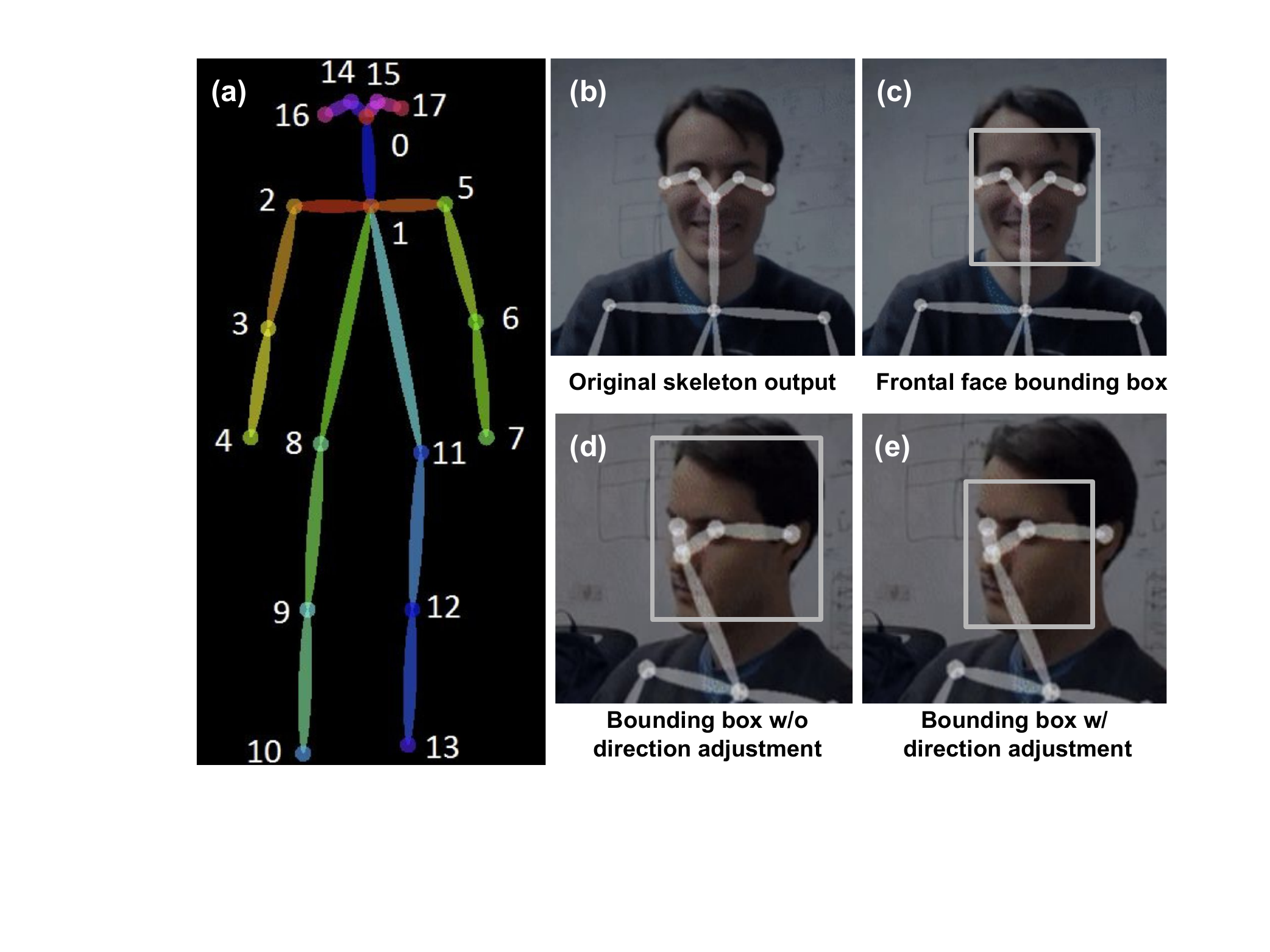}
\caption{\emph{(a) Sample OpenPose output. (b,c,d,e) \sys's approach adjusts the face's bounding box using the keypoints detected by OpenPose. (The face shown is selected from OpenPose project webpage~\cite{openpose})}}
\label{fig:pose_combined}
\end{figure}

\parab{Approach}
In this paper, we make the following observation: we can build end-to-end identity tracking using a state-of-the-art pose tracker. Specifically, we use, as a building block, a \textit{keypoint} based body pose tracker, called OpenPose \cite{openpose}. Given an image frame, OpenPose detects keypoints for each human in the image. Keypoints identify distinct anatomical structures in the body (\figref{fig:pose_combined}(a)) such as eyes, ears, nose, elbows, wrists, knees, hips \etc We can use these \textit{skeletons} for identification, tracking \textit{and} gesture recognition. OpenPose requires no pose calibration (unlike, say, the Kinect~\cite{bin2011study}), so it is attractive for our setting, and is fast, achieving up to 15 fps for body pose detection. (OpenPose also has modes where it can detect faces and hand gestures using many more keypoints than in \figref{fig:pose_combined}(a), but using these reduces the frame rate dramatically, and also has lower accuracy for shoppers far away from the camera).

However, fundamentally, since OpenPose operates only on a single frame, \sys needs to add identification, tracking \textit{and} gesture recognition algorithms on top of OpenPose to continuously identify and tracks shoppers and their gestures. The rest of this section describes these algorithms.

\parab{Shopper Identification}
\sys uses fast feature-based face recognition to identify shoppers. While prior work has explored other approaches to identification such as body features~\cite{bai2017scalable, chen2017beyond, zhao2017spindle} or clothing color \cite{lu2001color}, we use faces because (a) face recognition has been well-studied by vision researchers and we are likely to see continued improvements, (b) faces are more robust for identification than clothing color, and (c) face features have the highest accuracy in large datasets (\secref{sec:related}).

\parae{Feature-based face recognition}
When a user registers, \sys takes a video of their face, extracts features, and builds a fast classifier using these features. To identify shoppers, \sys does not directly use a face detector on the entire image because traditional HAAR based detectors~\cite{lienhart2002extended} can be inaccurate, and recent DNN-based face detectors such as MTCNN~\cite{zhang2016joint} can be slow. Instead, \sys \textit{identifies a face's bounding box using keypoints} from OpenPose, specifically, the five keypoints of the face from the nose, eyes, and ears (\figref{fig:pose_combined}(b)). Then, it extracts features from within the bounding box and applies the trained classifier.

\sys must (a) enable fast training of the classifier since this step is part of the registration process and registration is required to be fast (\secref{sec:identity}), (b) must robustly detect the bounding box for different facial orientations relative to the camera to avoid classification inaccuracy.

\parae{Fast Classification}
Registration is performed once for each customer. During registration, \sys extracts features from the customer's face. To do this, we evaluated several face feature extractors~\cite{baltruvsaitis2016openface, facefeature, schroff2015facenet}, and ultimately selected ResNet-34's feature extractor~\cite{facefeature} which produces a 128-dimension feature vector, performs best in both speed and accuracy (\secref{sec:choices}).

With these features, we can identify faces by comparing feature distances, build classifiers, or train a neural network. After experimenting with these options, we found that a $k$ nearest neighbor (kNN) classifier, in which each customer is trained as a new class, worked best among these choices (\secref{sec:choices}). \sys builds one kNN-based classifier for all customers and uses it across all cameras.

\parae{Tightening the face bounding box}
During normal operation, \sys extracts facial features after drawing a bounding box (derived from OpenPose keypoints) around each customer's face. \sys infers the face's bounding box width using the distance between two ears, and the height using the distance from nose to neck. This works well when the face points towards the camera (\figref{fig:pose_combined}(c)), but can result in an inaccurate bounding box when a customer faces slightly away from the camera (\figref{fig:pose_combined}(d)). This inaccuracy can degrade classification performance.

To obtain a tighter bounding box, we estimate head \textit{pitch} and \textit{yaw} using the keypoints. Consider the line between the nose and neck keypoints: the distance of each eye and ear keypoint to this axis can be used to estimate head \textit{yaw}. Similarly, the distance of the nose and neck keypoints to the axis between the ears can be used to estimate \textit{pitch}. Using these, we can tighten the bounding box significantly (\figref{fig:pose_combined}(e)). To improve detection accuracy (\secref{sec:eval}) when a customer's face is not fully visible in the camera, we also use face alignment~\cite{baltruvsaitis2016openface}, which estimates the frontal view of the face. 




\parab{Shopper Tracking}
A user's face may not always be visible in every frame, since customers may intentionally or otherwise turn their back to the camera. However, \sys needs to be able to identify the customer in frames where the customer's face is not visible, for which it uses \textit{tracking}. \sys assumes the use of existing security cameras, which, if placed correctly, make it unlikely that a customer can evade all cameras at all times (put another way, if the customer is able to do this, the security system's design is faulty). 

\begin{figure}
\centering
\includegraphics[width=0.9\columnwidth]{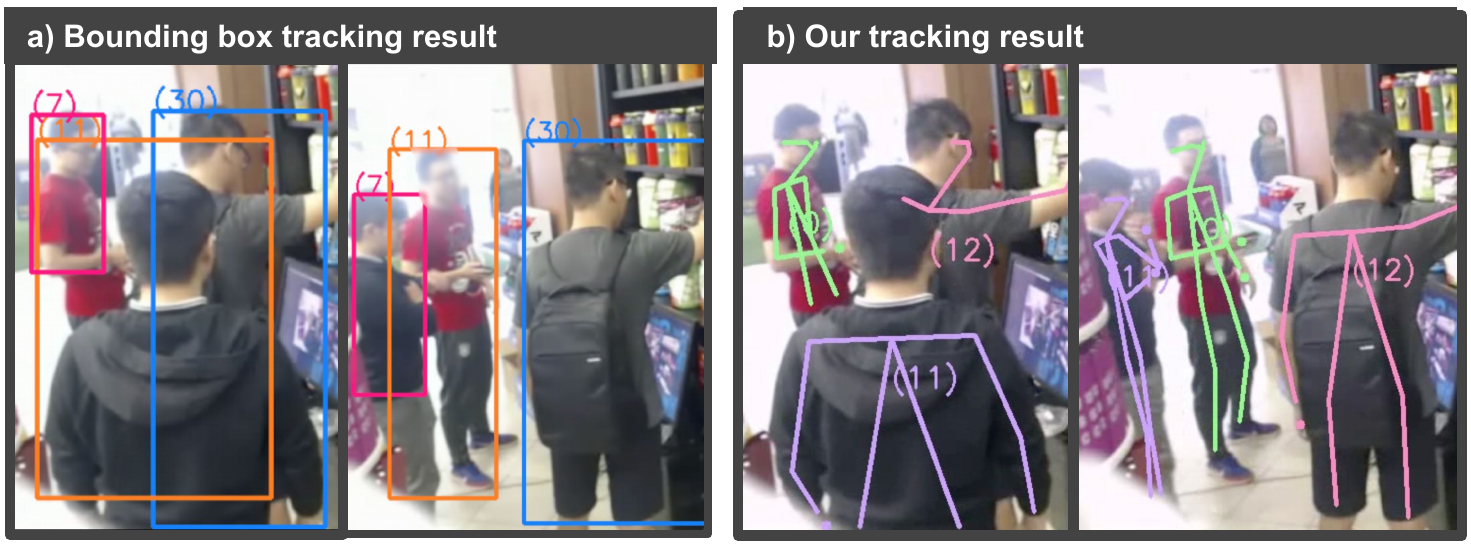}
\caption{\emph{Bounding-box based approaches (b) have trouble tracking multiple users in crowds, but our approach (a) works well in these settings.}}
\label{fig:tracking_res}
\end{figure}

\parae{Skeleton-based Tracking}
Existing human trackers use bounding box based approaches~\cite{milan2016mot16, leal2015motchallenge, tang2017multiple, wojke2017simpl, ristani2014tracking}, which can perform poorly in in-store settings with partial or complete occlusions (\figref{fig:tracking_res}(a)). We quantify this in \secref{sec:efficiency_eval} with the state-of-the-art bounding box based tracker, DeepSort~\cite{wojke2017simpl}, but \figref{fig:tracking_res} demonstrates this visually.

Instead, we use the skeleton generated by OpenPose to develop a tracker that uses geometric properties of the body frame. We use the term \textit{track} to denote the movements of a distinct customer (whose face may or may not have been identified). Suppose OpenPose identifies a skeleton in a frame: the goal of the tracker is to associate the skeleton with an existing track if possible. \sys uses the following to track customers. It tries to align each keypoint in the skeleton with the corresponding keypoint in the last seen skeleton in each track, and selects that track whose skeleton is the closest match (the sum of match errors is smallest). Also, as soon as it is able to identify the face, \sys associates the customer's identity with the track (to be robust to noise, \sys requires that the customer's face is identified in 3 successive frames). To work well, the tracking algorithm needs to correctly handle partial and complete occlusions.

\parae{Dealing with Partial Occlusions}
When a shopper's body is not completely visible (\eg because she is partially obscured by another customer, \figref{fig:tracking_res}(b)), OpenPose can only generate a subset of the key points. In this case, \sys matches only on the visible subset. However, with significant occlusions, very few key points may be visible. In this case, \sys attempts to  increase matching confidence using the color histogram of the visible upper body area. However, if the two matching approaches (color and skeletal) conflict with each other, \sys skips matching attempts until subsequent frames when this process is repeated.

\begin{figure}
\centering\includegraphics[width=0.8\columnwidth]{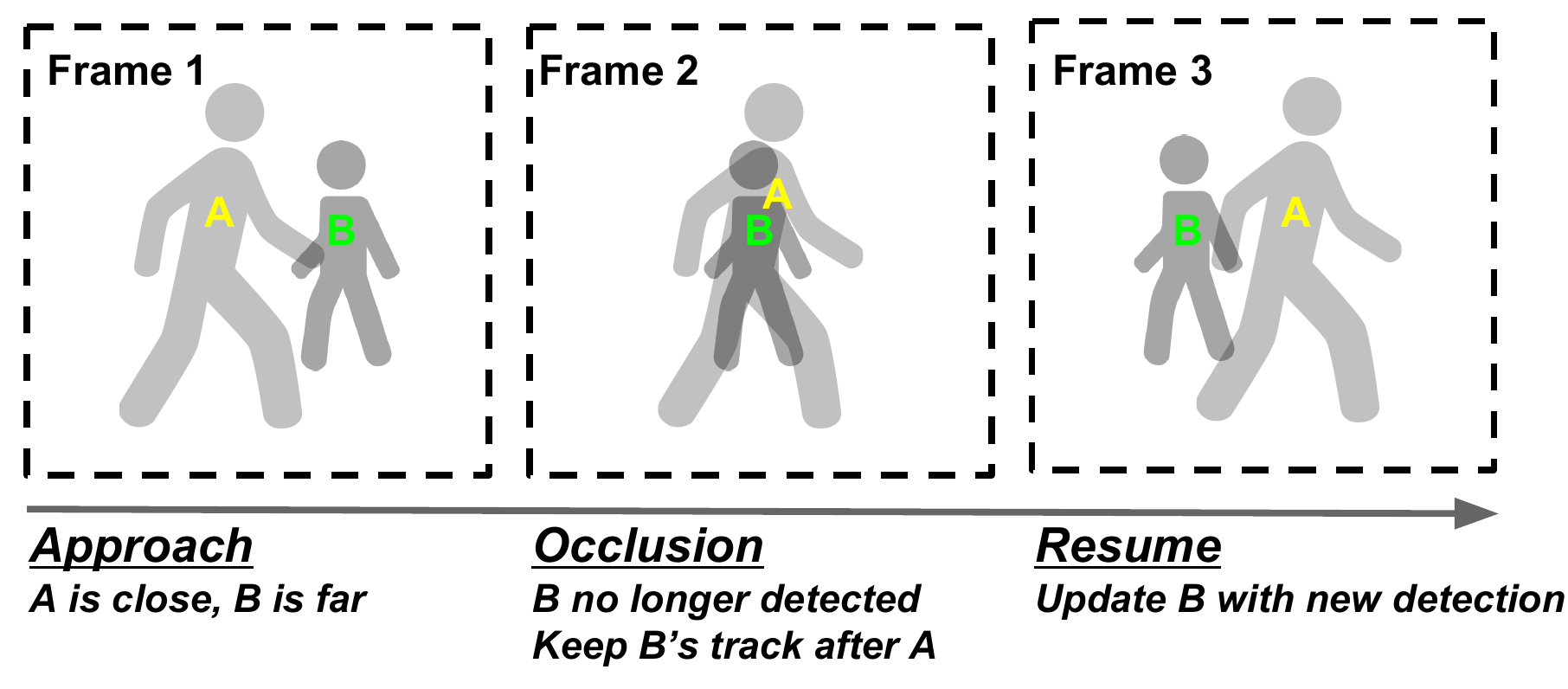}
\caption{\emph{When a shopper is occluded by another, \sys resumes tracking after the shopper re-appears in another frame (lazy tracking).}}
\label{fig:lazy_tracking}
\end{figure}

\parae{Dealing with Complete Occlusions}
In some cases, a shopper may be completely obscured by another. \sys uses lazy tracking (\figref{fig:lazy_tracking}) in this case. When an existing track disappears in the current frame, \sys checks if, in the previous frame, the track was close to the edge of the image, in which case it assumes the customer has moved out of the camera's field of view and deletes the track. Otherwise, it marks the track as \textit{blocked}. When the customer reappears in a subsequent frame, it reactivates the blocked track.

\begin{figure}
\centering
\includegraphics[width=0.9\columnwidth]{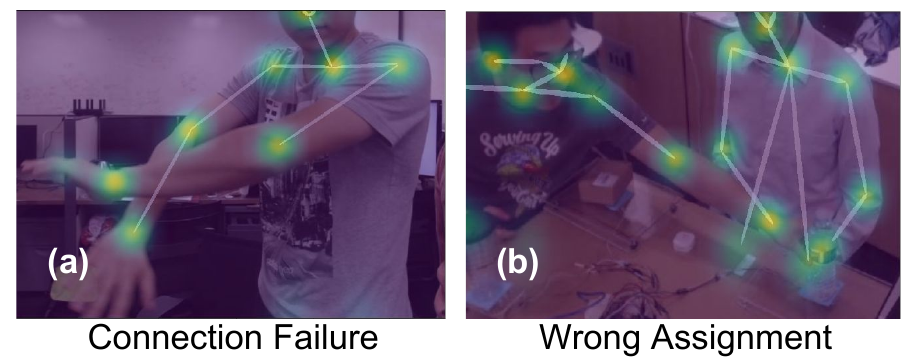}
\caption{\emph{OpenPose can (a) miss an assignment between elbow and wrist, or (b) wrongly assign one person's joint to another.}}
\label{fig:openposefail}
\end{figure}

\parab{Shopper Gesture Tracking}
\sys must recognize the arms of each shopper in order to determine which item he or she purchases (\secref{sec:behavior}). OpenPose has a built-in \textit{limb association} algorithm, which associates shoulder joints to elbows, and elbows to wrists. We have found that this algorithm is a little brittle in our setting: it can miss an association (\figref{fig:openposefail}(a)), or mis-associate part of a limb of one shopper with another (\figref{fig:openposefail}(b)). 

\parae{How limb association in OpenPose works}
OpenPose first uses a DNN to associate with each pixel confidence value of it being part of an anatomical key point (\eg an elbow, or a wrist). During image analysis, OpenPose also generates vector fields (called \textit{part affinity fields}~\cite{cao2017realtime}) for upper-arms and forearms whose vectors are aligned in the direction of the arm. Having generated keypoints, OpenPose then estimates, for each pair of keypoints, a measure of alignment between an arm's part affinity field, and the line between the keypoints (\eg elbow and wrist). It then uses a bipartite matching algorithm to associate the keypoints.

\parae{Improving limb association robustness}
One source of brittleness in OpenPose's limb association is the fact that the pixels for the wrist keypoint are conflated with pixels in the hand (\figref{fig:openposefail}(a)). This likely reduces the part affinity alignment, causing limb association to fail. To address this, for each keypoint, we filtered outlier pixels by removing pixels whose distance from the mediod~\cite{park2009simple} was greater than the 85th percentile.

\begin{figure}
\centering
\includegraphics[width=0.75\columnwidth]{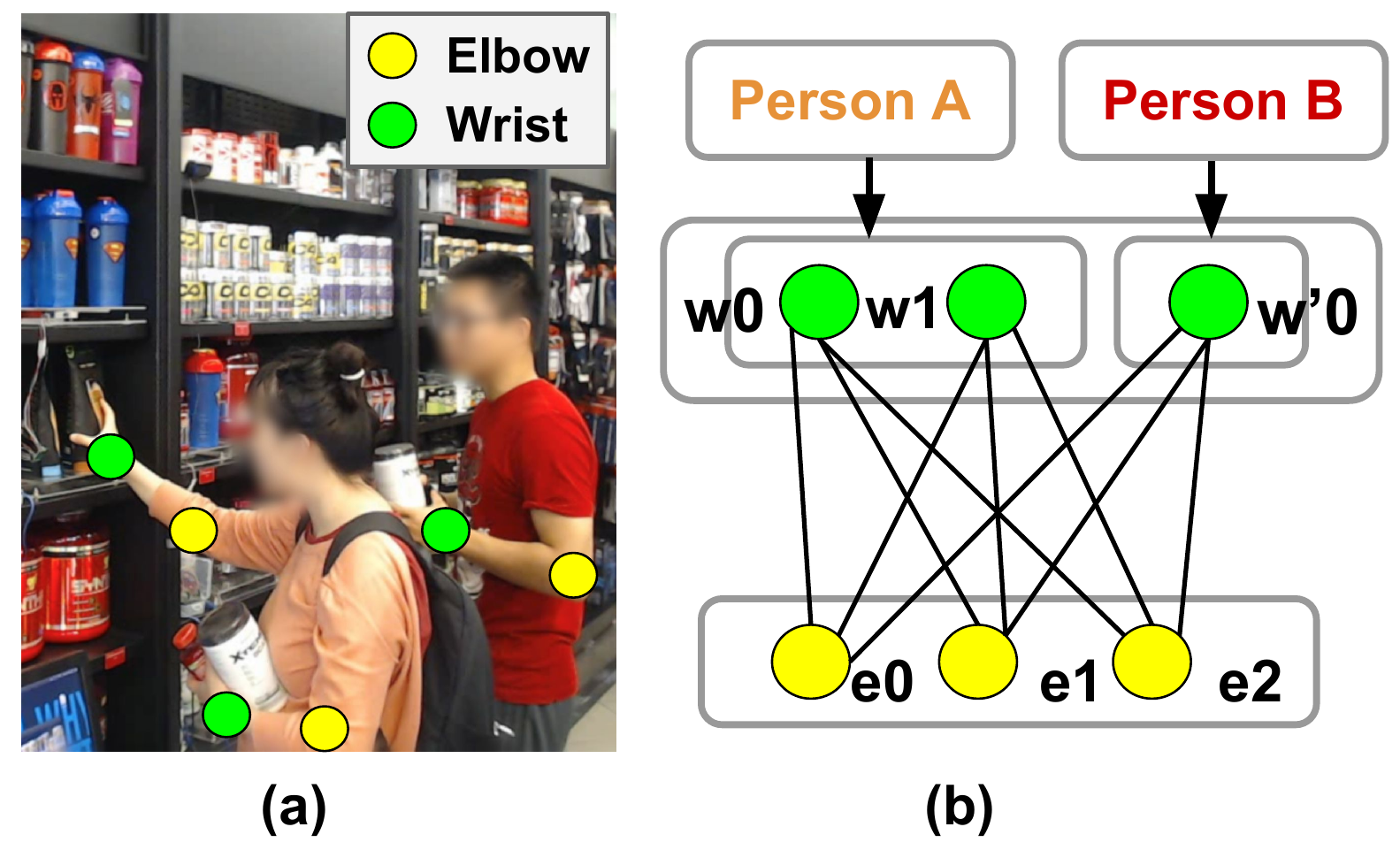}
\caption{\emph{OpenPose associates limb joints using bipartite matching.}} 
\label{fig:openposeopt}
\end{figure}

The second source of brittleness is that OpenPose's limb association treats each limb independently, resulting in cases where the key point from one person's elbow may get associated with another person's wrist (\figref{fig:openposefail}(b)). To avoid this failure mode, we modify OpenPose's limb association algorithm to treat one person's forearms or upper-arms as a pair (\figref{fig:openposeopt}). To identify forearms (or upper-arms) as belonging to the same person, we measure the Euclidean distance $ED(.)$ between color histograms $F(.)$ belonging to the two forearms, and treat them as a pair if the distance is less than an empirically-determined threshold $thresh$. Mathematically, we formulate this as an optimization problem:

\vspace{-2ex}
{\small
\begin{maxi*}{i,j}{\sum_{i \in E}\sum_{j \in W}{A_{i,j}z_{i,j}}}{}{}
   \addConstraint{\sum_{j \in W}{z_{i,j}}}{\leq 1\quad}{\forall i \in E}
   \addConstraint{\sum_{i \in E}{z_{i,j}}}{\leq 1\quad}{\forall j \in W}
   \addConstraint{ED(F(i,j),F(i',j'))}{< thresh\quad}{\forall j, j' \in W \ i, i' \in E}
\end{maxi*}
}
\vspace{-2ex}

where $E$ and $W$ are the sets of elbow and wrist joints, and $A_{i,j}$ is the alignment measure between the $i$-th elbow and the $j$-th wrist, while $z_{i,j}$ is an indicator variable indicating connectivity between the elbow and the wrist. The third constraint models whether two elbows belong to the same body, using the Euclidean distance between the color histograms of the body color. This formulation reduces to a max-weight bipartite matching problem, and we solve it with the Hungarian algorithm~\cite{kuhn1955hungarian}.

\begin{figure}[htbp]
\centering
\includegraphics[width=0.85\columnwidth]{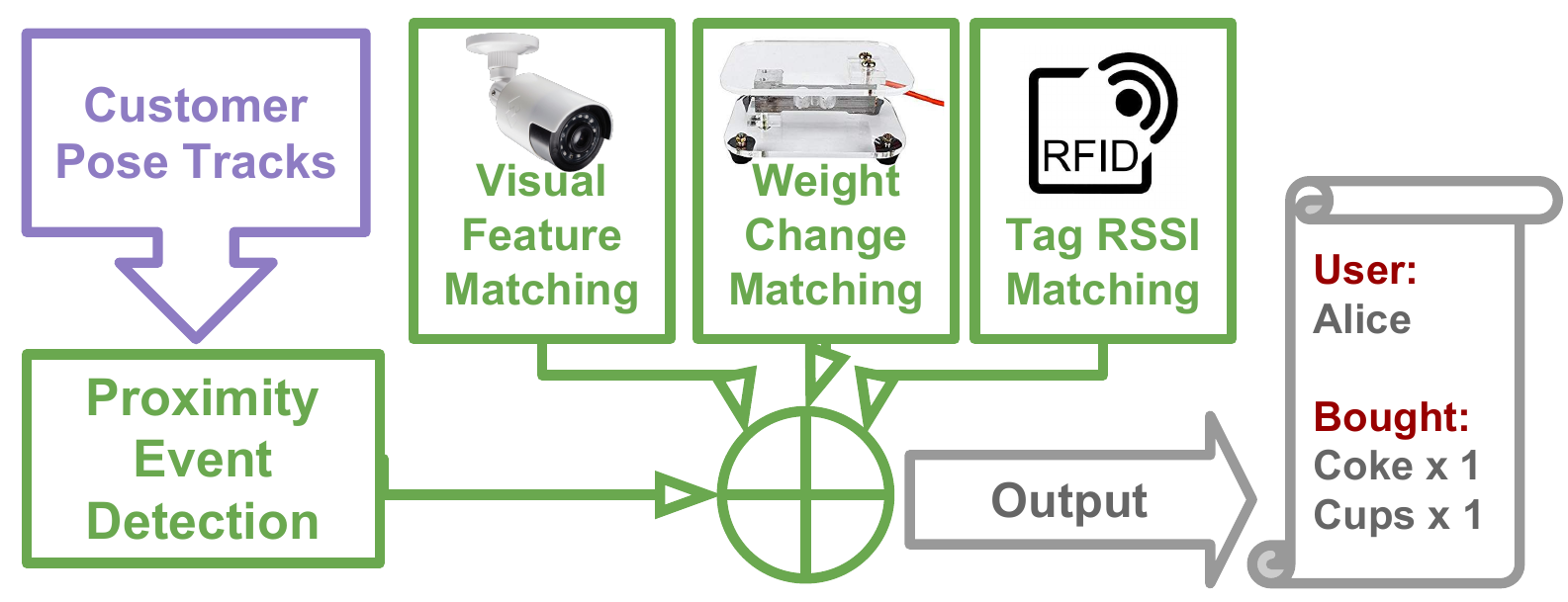}
\caption{\textit{\sys recognizes the items a shopper picks up by fusing vision with smart-shelf sensors including weight and RFID.}}
\label{fig:behavior_workflow}
\end{figure}

\subsection{Shopper Action Recognition}
\label{sec:behavior}

When a shopper is being continuously tracked, and their hand movements accurately detected, the next step is to recognize hand actions, specifically to identify item(s) which the shopper picks up from a shelf. Vision-based hand tracking alone is insufficient for this in the presence of multiple shoppers concurrently accessing items under variable lighting conditions. \sys leverages the fact that many retailers are installing smart shelves~\cite{smart_shelves,smart_shelves2} to deter theft. These shelves have weight sensors and are equipped with RFID readers. Weight sensors cannot distinguish between items of similar weight, while not all items are likely to have RFID tags for cost reasons. So, rather than relying on any individual sensor, \sys fuses detections from cameras, weight sensors, and RFID tags to recognize hand actions.

\parab{Modeling the sensor fusion problem}
In a given camera view, at any instant, multiple shoppers might be reaching out to pick items from shelves. Our identity tracker (\secref{sec:identity}) tracks hand movement, the goal of the action recognition problem is to associate each shopper's hand with the item he or she picked up from the shelf. We model this association between shopper's hand $k$ and item $m$ as a probability $p_{k,m}$ derived from fusing cameras, weight sensors, and RFID tags (\figref{fig:behavior_workflow}). $p_{k,m}$ is itself derived from \textit{association probabilities} for each of the devices, in a manner described below. Given these probabilities, we then solve the association problem using a maximum weight bipartite matching. In the following paragraphs, we discuss details of each of these steps.

\parab{Proximity event detection}
Before determining association probabilities, we need to determine when a shopper's hand approaches a shelf. This proximity event is determined using the identity tracker module's gesture tracking (\secref{sec:identity}). Knowing where the hand is, \sys uses image analysis to determine when a hand is close to a shelf. For this, \sys requires an initial configuration step, where store administrators specify camera view parameters (mounting height, field of view, resolution \etc), and which shelf/shelves are where in the camera view. \sys uses a threshold pixel distance from hand to the shelf to define proximity, and its identity tracker reports \textit{start} and \textit{finish} times for when each hand is within the proximity of a given shelf (a \textit{proximity event}).

\begin{figure}[t]
\centering\includegraphics[width=0.8\columnwidth]{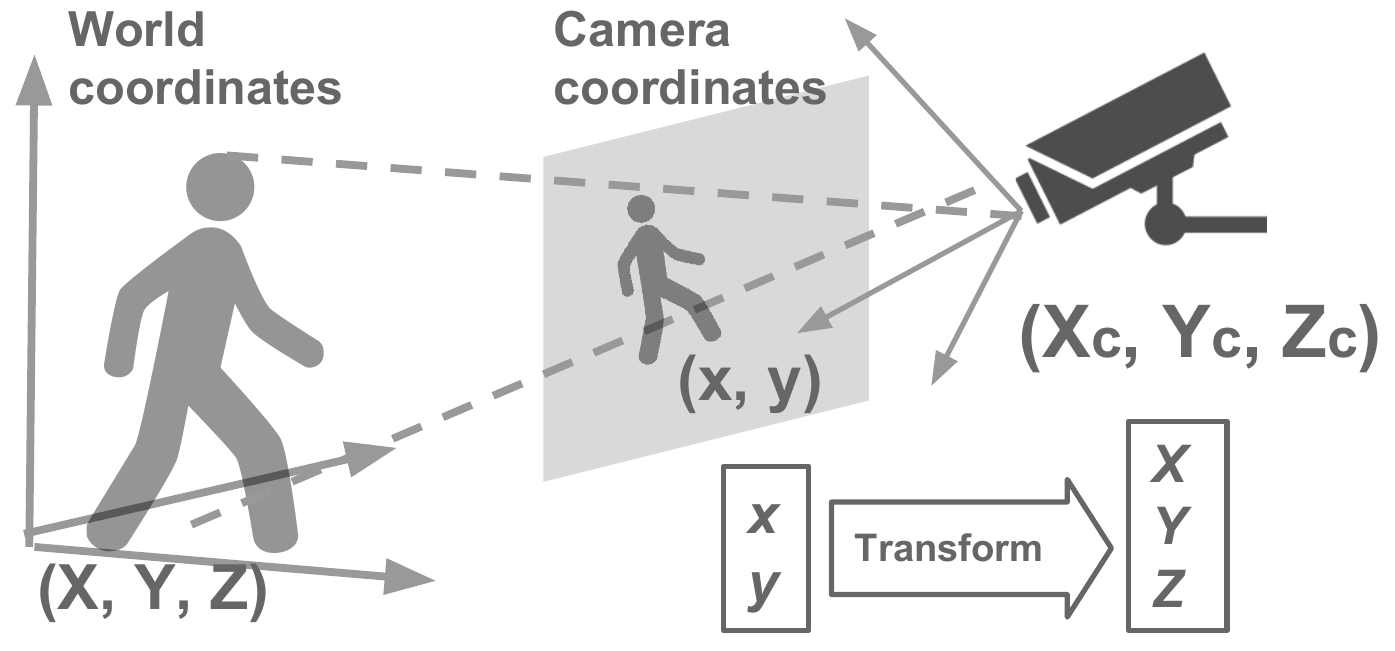}
\caption{\emph{When the shopper's hand is obscured, \sys infers proximity to shelves by determining when a shoppers ankle joint is near a shelf.}}
\label{fig:2d_mapping}
\end{figure}

In some cases, the hand may not be visible. In these cases, \sys estimates proximity using the skeletal keypoints identified by OpenPose (\secref{sec:identity}). Specifically, \sys knows, from the initial configuration step, the camera position (including its height), its orientation, and its field of view. From this, and simple geometry, it can estimate the pixel position of any point on the visible floor. In particular, it can estimate the pixel location of a shopper's ankle joint (\figref{fig:2d_mapping}), and use this to estimate the distance to a shelf. When the ankle joint is occluded, we extrapolate its position from the visible part of the skeleton to estimate the position.

\begin{figure*}[htbp]
   \begin{minipage}{0.32\linewidth}
    \centerline{\includegraphics[width=0.85\columnwidth]{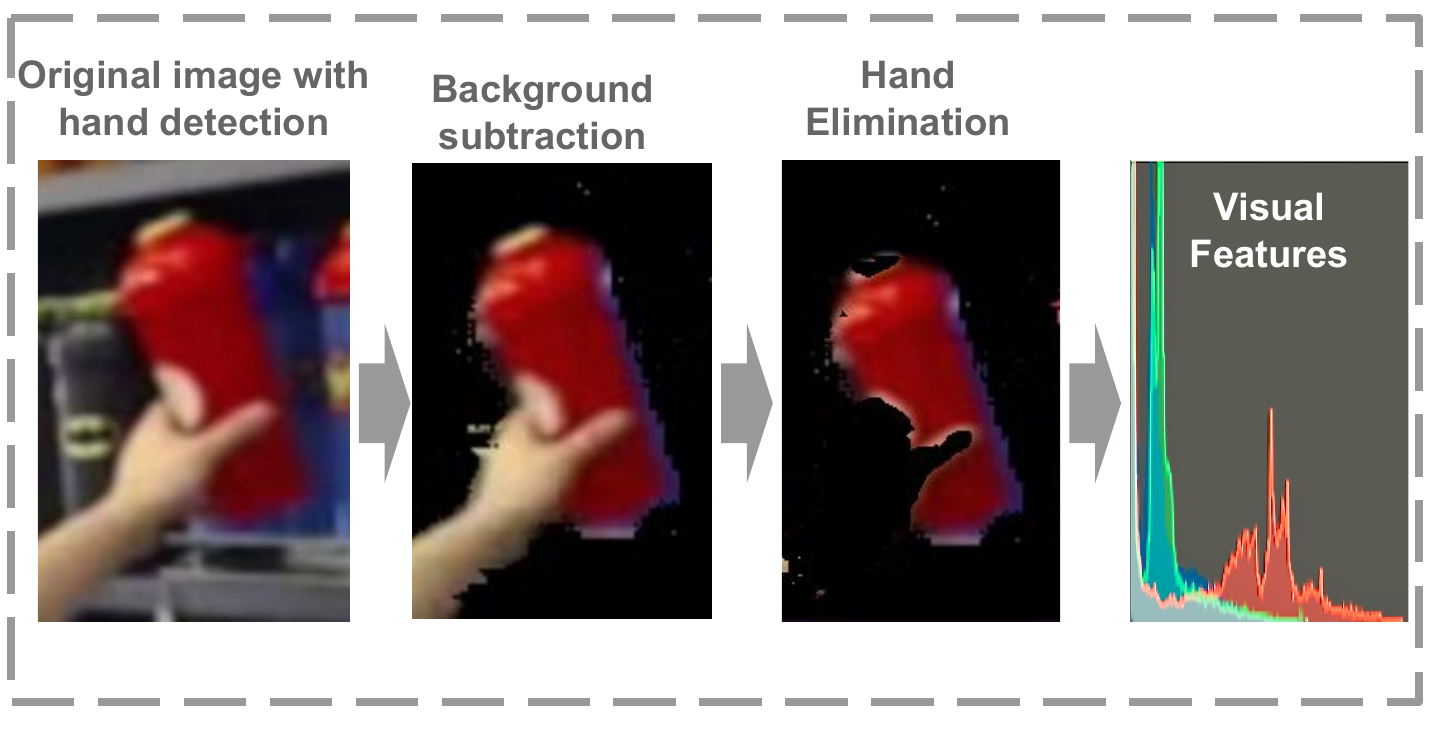}}
    \centerline{(a)}
    \label{fig:vision_matching}
  \end{minipage}
  \begin{minipage}{0.32\linewidth}
    \centerline{\includegraphics[width=0.85\columnwidth]{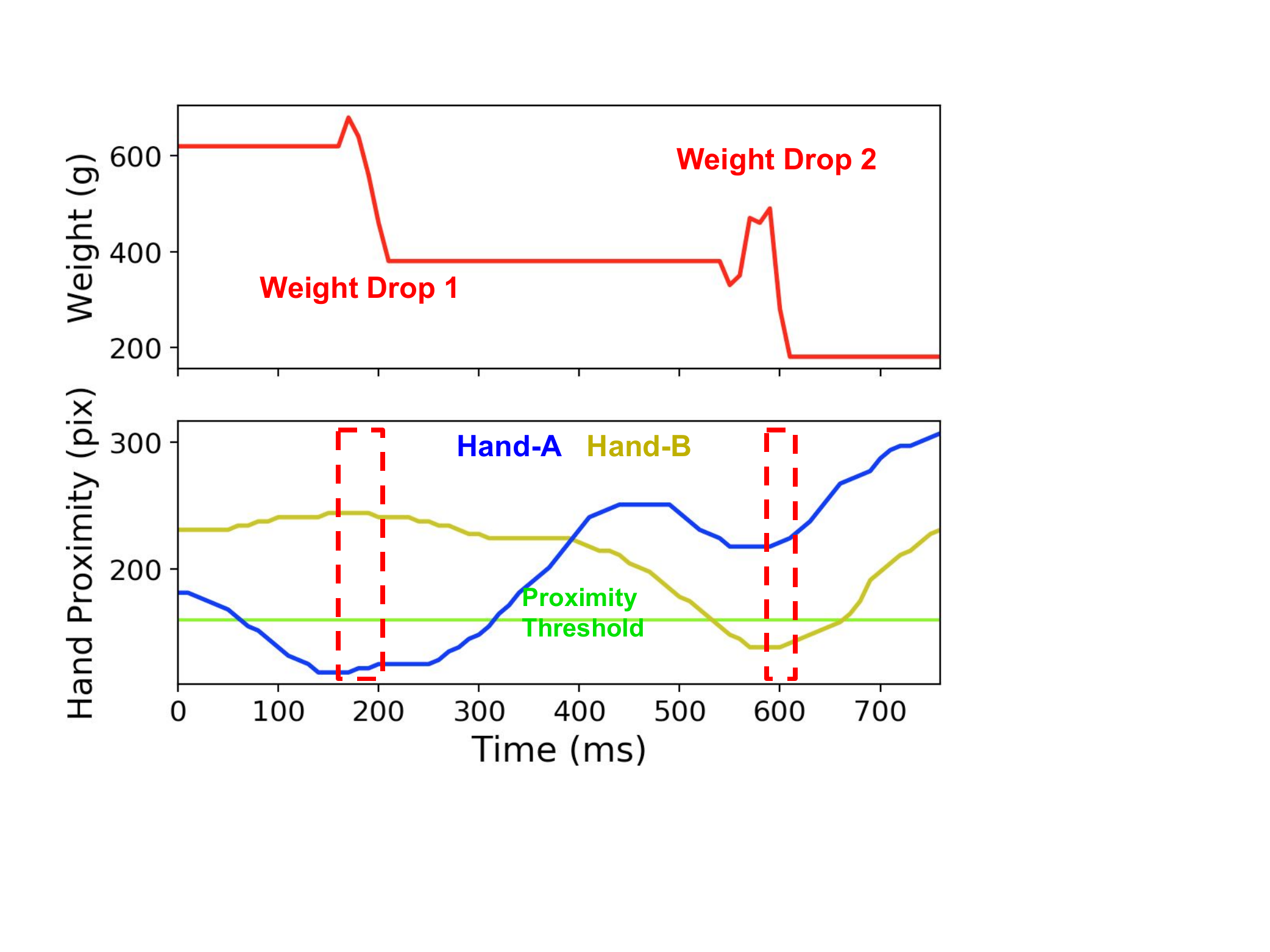}}
    \centerline{(b)}
    \label{fig:weight_matching}
  \end{minipage}
    \begin{minipage}{0.32\linewidth}
    \centerline{\includegraphics[width=0.85\columnwidth]{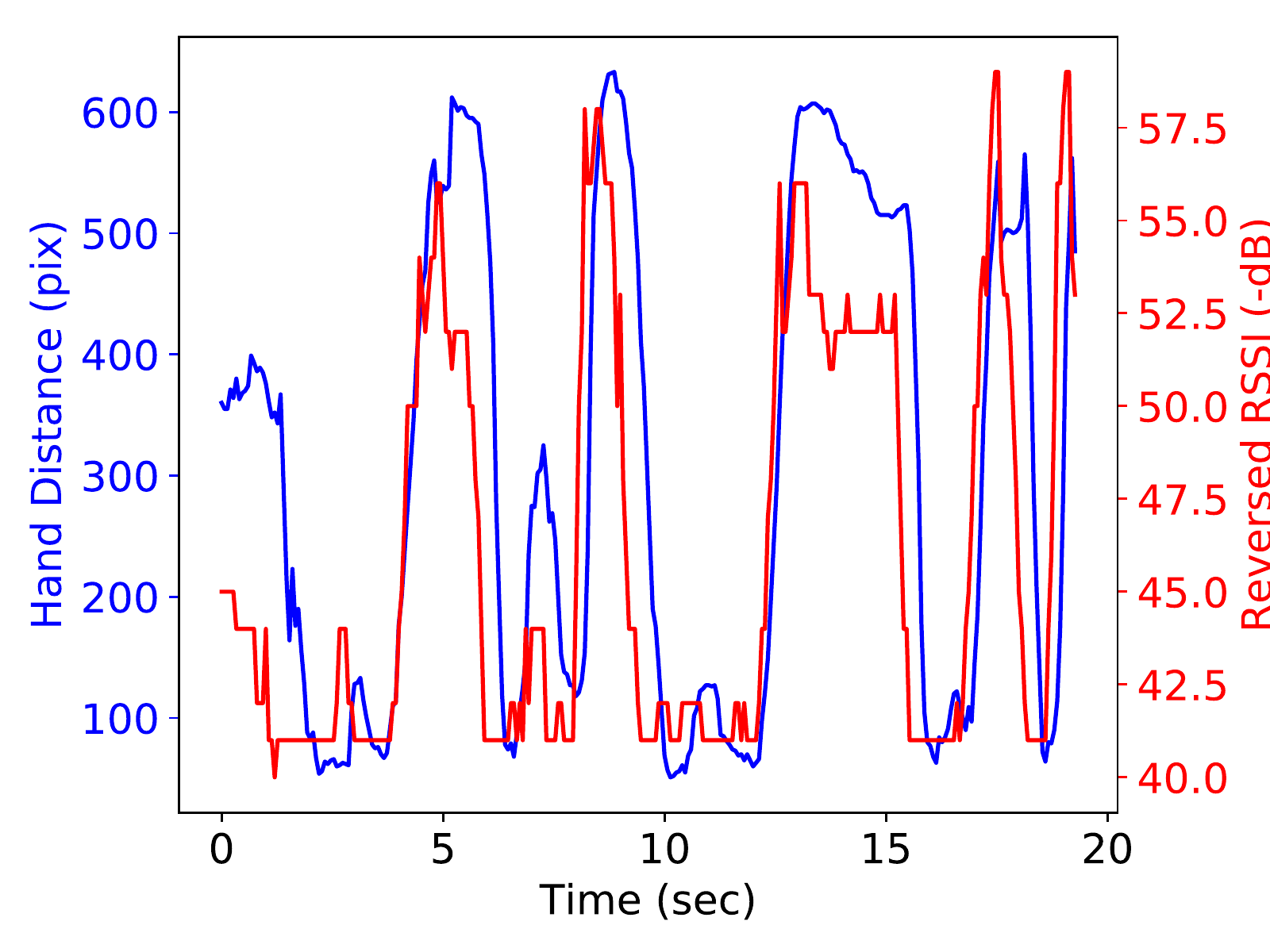}}
    \centerline{(c)}
    \label{fig:rssi_distance}
  \end{minipage}
  \vspace{-2pt}
  \caption{\emph{(a) Vision based item detection does background subtraction and removes the hand outline. (b) Weight sensor readings are correlated with hand proximity events to assign association probabilities. (c) Tag RSSI and hand movements are correlated, which helps associate proximity events to tagged items.}}
  \label{fig:behavior_figs}
\end{figure*}

\parab{Association probabilities from the camera}
When a proximity event starts, \sys starts tracking the hand and any item in the hand. It uses the color histogram of the item to classify the item. To ensure robust classification, \sys performs (\figref{fig:behavior_figs}(a)) (a) background subtraction to remove other items that may be visible and (b) eliminates the hand itself from the item by filtering out pixels whose color matches typical skin colors. \sys extracts a 384 dimension color histogram from the remaining pixels.

During an initial configuration step, \sys requires store administrators to specify which objects are on which shelves. \sys then builds, for each shelf (a single shelf might contain 10-15 different types of items), builds a feature-based kNN classifier (chosen both for speed and accuracy). Then, during actual operation, when an item is detected, \sys runs this classifier on its features. The classifier outputs an ordered list of matching items, with associated match probabilities. \sys uses these as the association probabilities from the camera. Thus, for each hand $i$ and each item $j$, \sys outputs the camera-based association probability.

\parab{Association probabilities from weight sensors} 
In principle, a weight sensor can determine the reduction in total weight when an item is removed from the shelf. Then, knowing which shopper's hand was closest to the shelf, we can associate the shopper with the item. In practice, this association needs to consider real-world behaviors. First, if two shoppers concurrently remove two items of different weights (say a can of Pepsi and a peanut butter jar), the algorithm must be able to identify which shopper took which item. Second, if two shoppers are near the shelf, and two cans of Pepsi were removed, the algorithm must be able to determine if a single shopper took both, or each shopper took one. To increase robustness to these, \sys breaks this problem down into two steps: (a) it associates a proximity event to \textit{dynamics in scale readings}, and (b) then associates scale dynamics to items by detecting weight changes.

\parae{Associating proximity events to scale dynamics}
Weight scales sample readings at 30 Hz. At these rates, we have observed that, when a shopper picks up an item or deposits an item on a shelf, there is a distinct "bounce" (a peak when an item is added, or a trough when removed) because of inertia (\figref{fig:behavior_figs}(b)). If $d$ is the duration of this peak or trough, and $d'$ is the duration of the proximity event, we determine the association probability between the proximity event and the peak or trough as the ratio of the intersection of the two to the union of the two. As \figref{fig:behavior_figs}(b) shows, if two shoppers pick up items at almost the same time, our algorithm is able to distinguish between them. Moreover, to prevent shoppers from attempting to confuse \sys by temporarily activating the weight scale with a finger or hand, \sys filters out scale dynamics where there is high frequency of weight change.

\parae{Associating scale dynamics to items}
The next challenge is to measure the weight of the item removed or deposited. Even when there are multiple concurrent events, the 30 Hz sampling rate ensures that the peaks and troughs of two concurrent actions are likely distinguishable (as in \figref{fig:behavior_figs}(b)). In this case, we can estimate the weight of each item from the sensor reading at the beginning of the peak or trough $w_s$ and the reading at the end $w_e$. Thus $|w_s-w_e|$ is an estimate of the item weight $w$. Now, from the configuration phase, we know the weights of each type of item on the shelf. Define $\delta_j$ as $|w-w_j|$ where $w_j$ is the known weight of the $j$-th type of item in the shelf. Then, we say that the probability that the item removed or deposited was the $j$-th item is given by $\frac{1/{\delta}_{j}}{\sum_i (1/{\delta}_{i})}$. This definition accounts for noise in the scale (the estimates for $w$ might be slightly off) and for the fact that some items may be very similar in weight.

\parae{Combining these association probabilities} 
From these steps, we get two association probabilities: one associating a proximity event to a peak or trough, another associating the peak or trough to an item type. \sys multiplies these two to get the probability, according to the weight sensor, that hand $i$ picked item $j$.

\parab{Association probabilities from RFID tag}
For items which have an RFID tag, it is trivial to determine which item was taken (unlike with weight or vision sensors), but it is still challenging to associate proximity events with the corresponding items. For this, we leverage the fact that the tag's RSSI becomes weaker as it moves away from the RFID reader. \figref{fig:behavior_figs}(c) illustrates an experiment where we moved an item repeatedly closer and further away from a reader; notice how the changes in the RSSI closely match the distance to the reader. In smart shelves, the RFID reader is mounted on the back of the shelf, so that when an object is removed, its tag's RSSI decreases. To determine the probability that a given hand caused this decrease, we use probability-based Dynamic Time Warping~\cite{bautista2013probability}, which matches the time series of hand movements with the RSSI time series and assigns a probability which measures the likelihood of association between the two. We use this as the association probability derived from the RFID tag.




\parab{Putting it all together} In the last step, \sys  formulates an assignment problem to determine which hand to associate with which item. First, it determines a time window consisting of a set of overlapping proximity events. Over this window, it first uses the association probabilities from each sensor to define a composite probability $p_{k,m}$ between the $k$-th hand and the $m$-th item: $p_{k,m}$ is a weighted sum of the three probabilities from each sensor (described above), with the weights being empirically determined.

Then, \sys formulates the assignment problem as an optimization problem:

\vspace{-2ex}
{\small
\begin{maxi*}
  {k,m}{\sum{p_{k,m}z_{k,m}}}{}{}
  \addConstraint{\sum_{k \in H}{z_{k,m}}}{\leq 1\quad}{\forall m \in I}
  \addConstraint{\sum_{l \in I_t}{z_{k,l}}}{\leq u_l}{\forall k \in H}
\end{maxi*}
}
\vspace{-1ex}

\noindent
where $H$ is the set of hands, $I$ is the set of items, and $I_t$ is the set of \textit{item types}, and $z_{k,m}$ is an indicator variable that determines if hand $k$ picked up item $m$. The first constraint models the fact that each item can be removed or deposited by one hand, and the second models the fact that sometimes shoppers can pick up more than one item with a single hand: $u_l$ is a statically determined upper bound on the number of items of the $l$-th item that a shopper can pick up using a single hand (\eg it may be physically impossible to pick up more than 3 bottles of a specific type of shampoo). This formulation is a max-weight bipartite matching problem, which we can optimally solve using the Hungarian~\cite{kuhn1955hungarian} algorithm.

\subsection{GPU Multiplexing} 
\label{sec:scale}

Because retailer margins can be small, \sys needs to minimize overall costs. The computing infrastructure (specifically, GPUs) is an important component of this cost. In what we have described so far, each camera in the store needs a GPU. 

\begin{figure}
\centering
\includegraphics[width=0.9\columnwidth]{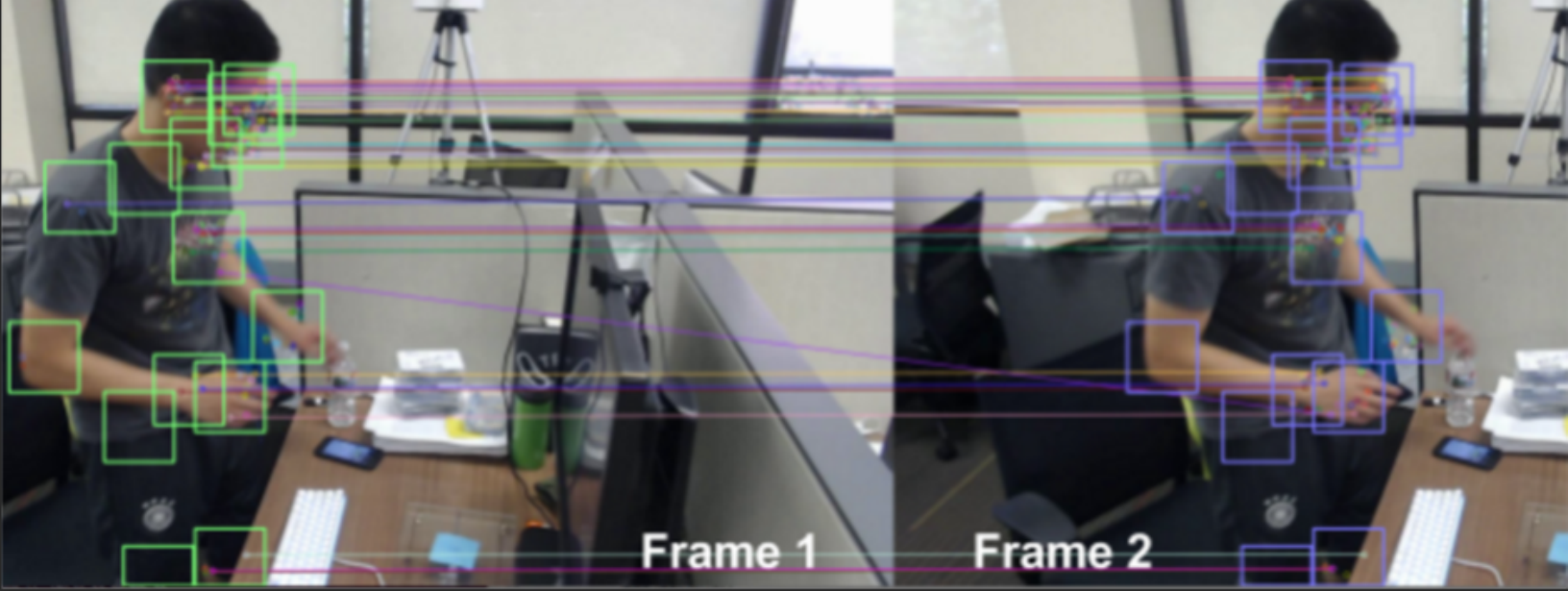}
\caption{\emph{ORB features from each joint bounding box are tracked across successive frames to permit multiplexing the GPU across multiple cameras.}}
\label{fig:joint_tracking}
\end{figure}

\sys actually enables multiple cameras to be multiplexed on one GPU. It does this by
avoiding running OpenPose on every frame. Instead, \sys uses a \textit{tracker} to track joint positions from frame to frame: these tracking algorithms are fast and do not require the use of the GPU. Specifically, suppose \sys runs OpenPose on frame $i$. On that frame, it computes ORB~\cite{ORB} features around every joint (\figref{fig:joint_tracking}(a)): ORB features can be computed faster than previously proposed features like SIFT and SURF. Then, for each joint, it identifies the position of the joint in frame $i+1$ by matching ORB features between the two frames. Using this it can reconstruct the skeleton in frame $i+1$ without running OpenPose on that frame.

\sys uses this to multiplex a GPU over $N$ different cameras. It runs OpenPose from a frame on each camera in a round-robin fashion. If a frame has been generated by the $k$-the camera, but \sys is processing a frame from another (say, the $m$-th) camera, then \sys runs feature-based tracking on the frame from the $k$ camera. Using this technique, we show that \sys is able to scale to using 4 cameras on one GPU without significant loss of accuracy (\secref{sec:eval}).

\section{Evaluation}
\label{sec:eval}

We now evaluate the end-to-end accuracy of \sys and explre the impact of each of our optimizations on overall performance.~\footnote{Demo video of \sys: https://vimeo.com/245274192}

\subsection{\sys Implementation}

\sepfootnotecontent{hx711}{The HX711 can sample at 80 Hz, but the Arduino MCU, when used with several weight scales, limits the sampling rate to 30 Hz.}

\begin{figure}[t]
  \begin{minipage}{0.49\linewidth}
    \centerline{\includegraphics[width=0.95\columnwidth]{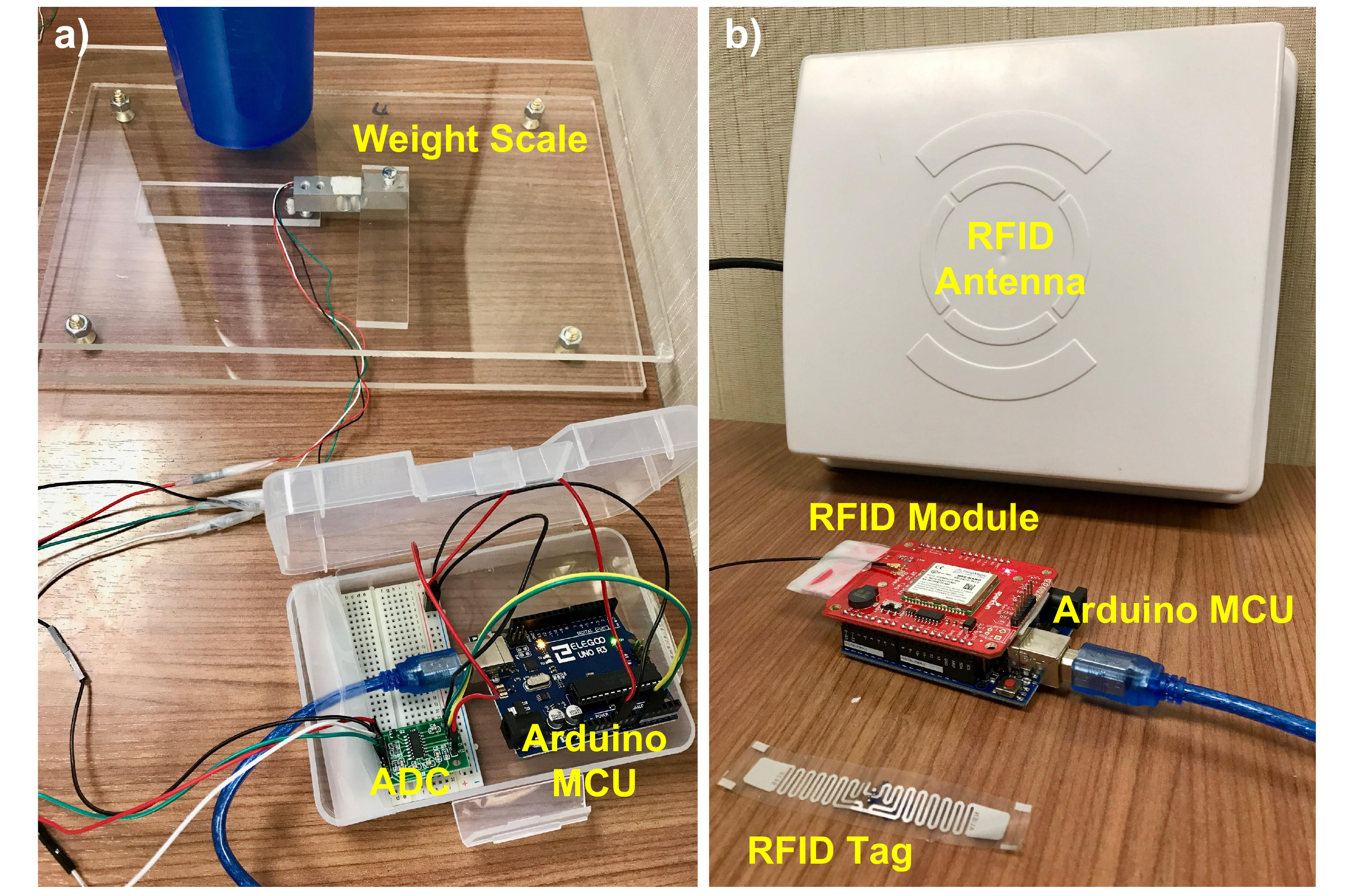}}
    \centerline{(a)}
    \label{fig:sys_demo}
  \end{minipage}
    \begin{minipage}{0.49\linewidth}
    \centerline{\includegraphics[width=0.95\columnwidth]{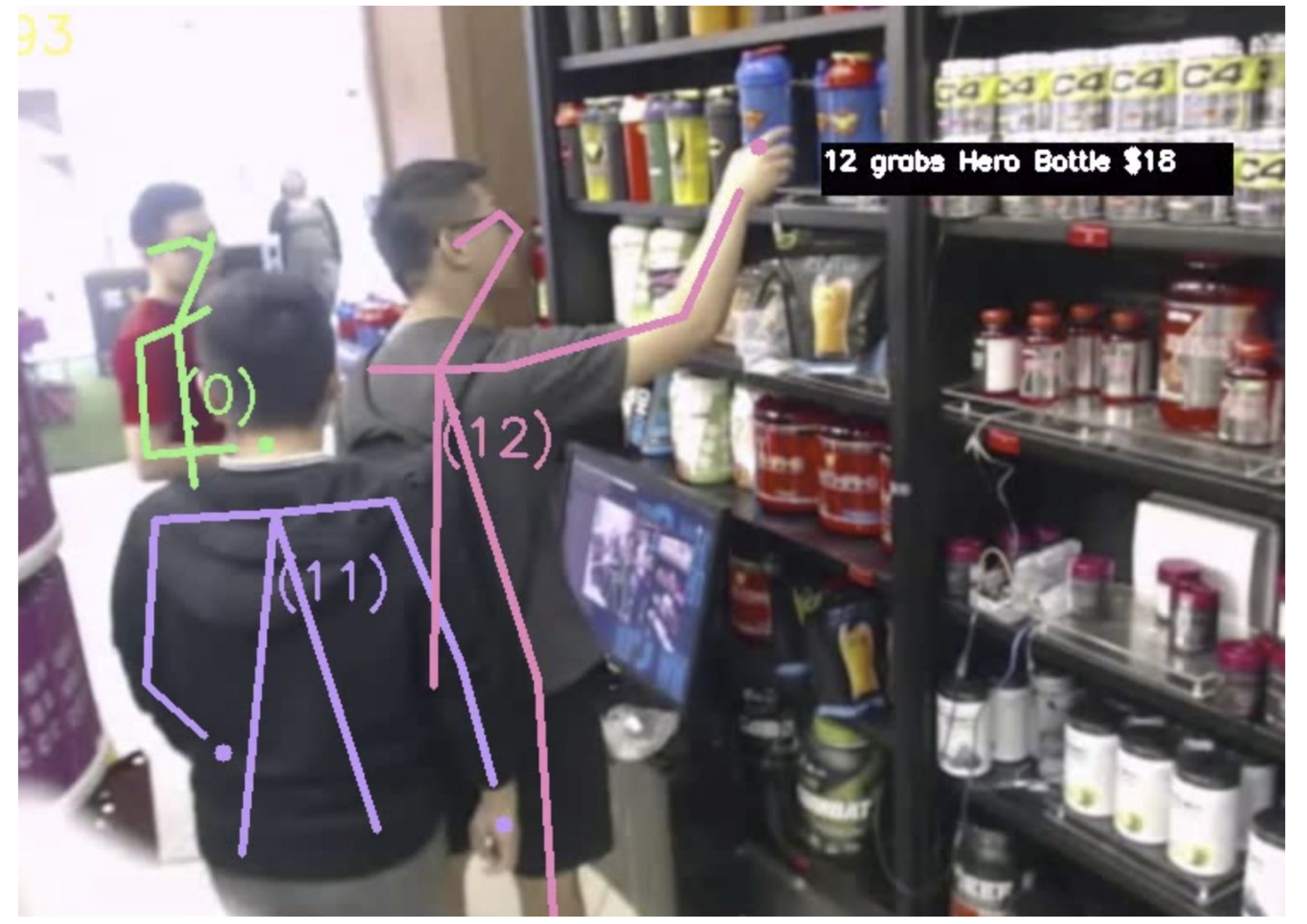}}
    \centerline{(b)}
    \label{fig:tracking_eva}
  \end{minipage}
  \caption{\emph{(a) Left: Weight sensor hardware, Right: RFID hardware; (b) \sys sample output.}}
  \label{fig:eva_figs}
\end{figure}

\parab{Weight-sensing Module}
To mimic weight scales on smart shelves, we built scales costing \$6, with fiberglass boards and 2~kg, 3~kg, 5G~kg pressure sensors. The sensor output is converted by the SparkFun HX711 load cell amplifier~\cite{hx711} to digital serial signals. An Arduino Uno Micro Control Unit (MCU)~\cite{uno} (\figref{fig:eva_figs}(a)-left) batches data from the ADCs and sends it to a server. The MCU has nine sets of serial Tx and Rx so it can collect data from up to nine sensors simultaneously. The sensors have a precision of around 5\-10~g, with an effective sampling rate of 30~Hz\sepfootnote{hx711}.  

\parab{RFID-sensing Module}
For RFID, we use the SparkFun RFID modules with antennas and multiple UHF passive RFID tags~\cite{sparkfun} (\figref{fig:eva_figs}(a)-right). The module can read up to 150 tags per second and its maximum detection range is 4~m with and antenna. The RFID module interfaces with the Arduino MCU to read data from tags.

\parab{Video input}
We use IP cameras~\cite{ipcam} for video recording. In our experiments, the cameras are mounted on merchandise shelves and they stream 720p video using Ethernet. We also tried webcams and they achieved similar performance (detection recall and precision) as IP cameras.

\parab{Identity tracking and action recognition}
These modules are built on top of the OpenPose~\cite{openpose} library's skeleton detection algorithm. As discussed earlier, we use a modified limb association algorithm. Our other algorithms are implemented in Python, and interface with OpenPose using a boost.python wrapper. Our implementation has over 4K lines of code.

\subsection{Methodology, Metrics, and Datasets}
\label{sec:meth}

\parab{In-store deployment}
To evaluate \sys, we collected traces from an actual deployment in a retail store. For this trace collection, we installed the sensors described above in two shelves in the store. First, we placed two cameras at the ends of an aisle so that they could capture both the people's pose and the items on the shelves. Then, we installed weight scales on each shelf. Each shelf contains multiple types of items, and all instances of a single item were placed on a single shelf at the beginning of the experiment (during the experiment, we asked users to move items from one shelf to another to try to confuse the system, see below). In total, our shelves contained 19 different types of items. Finally, we placed the RFID reader's antenna behind the shelf, and we attached RFID tags to all instances of 8 types of items.

\parab{Trace collection}
We then recorded five hours worth of sensor data from 41 users who registered their faces with \sys. We asked these shoppers to test the system in whatever way they wished to (\figref{fig:eva_figs}(b)). The shoppers selected from among the 19 different types of items, and interacted with the items (either removing or depositing them) a total of 307 times. Our cameras saw an average of 2.1 shoppers and a maximum of 8 shoppers in a given frame. In total, we collected over 10GB of video and sensor data, using which we analyze \sys' performance. 

\parab{Adversarial actions}
During the experiment, we also asked shoppers to perform three kinds of \textit{adversarial actions}. (1) \emph{Item-switching}: The shopper takes two items of similar color or similar weight and then puts one back, or takes one item and puts it on a different scale; (2) \emph{Hand-hiding}: The shopper hides the hand from the camera and grabs the item; (3) \emph{Sensor-tampering}: The shopper presses the weight scale with their hand. Of the 307 recorded actions, nearly 40\% were adversarial: 53 item-switching, 34 hand-hiding, and 31 sensor-tampering actions.


\parab{Metrics}
To evaluate \sys's accuracy, we use \textit{precision} and \textit{recall}. In our context, precision is the ratio of true positives to the sum of true positives and false positives.  Recall is the ratio of true positives to the sum of true positives and false negatives. For example, suppose a shopper picks items A, B, and C, but \sys shows that she picks items A, B, D, and E. A and B are correctly detected so the true positives are 2, but C is missing and is a false negative. The customer is wrongly associated with D and E so there are 2 false positives. In this example, recall is 2/3 and precision is 2/4.

\subsection{Accuracy of \sys}
\label{sec:accuracy_eval}

\begin{figure*}[t]
   \begin{minipage}{0.99\linewidth}
    \centerline{\includegraphics[width=0.85\columnwidth]{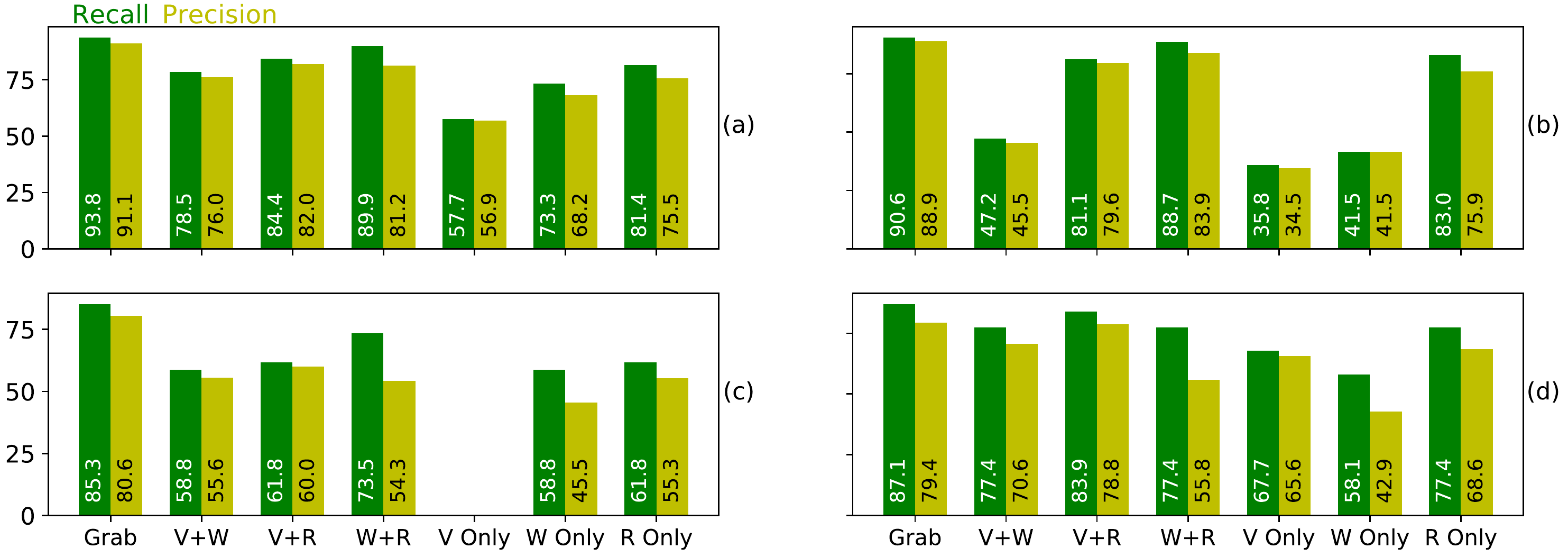}} 
  \end{minipage}
  \caption{\emph{\sys has high precision and recall across our entire trace (a), relative to other alternatives that only use a subset of sensors (\emph{W: Weight; V: Vision; R: RFID}), even under adversarial actions such as (b) Item-switching; (c) Hand-hiding; (d) Sensor-Tampering .}}
  \label{fig:action_accuracy}
\end{figure*}

\parab{Overall precision and recall}
\figref{fig:action_accuracy}(a) shows the precision and recall of \sys, and quantifies the impact of using different combinations of sensors: using vision only (\textit{V Only}), weight only (\textit{W only}), RFID only (\textit{R only}) or all possible combinations of two of these sensors. Across our entire trace, \sys achieves a recall of nearly 94\% and a precision of over 91\%. This is remarkable, because in our dataset nearly 40\% of the actions are adversarial (\secref{sec:meth}). We dissect \sys failures below and show how these are within the loss margins that retailers face today due to theft or faulty equipment.

\sepfootnotecontent{singsens}{For computing the association probabilities \secref{sec:behavior}. Cameras are still used for identity tracking and proximity event detection.}

\sepfootnotecontent{rfid}{In general, since RFID is expensive, not all objects in a store will have RFID tags. In our deployment, a little less than half of the item types were tagged, and these numbers are calculated only for tagged items.}

Using only a single sensor\sepfootnote{singsens} degrades recall by 12-37\% and precision by 16-36\% (\figref{fig:action_accuracy}(a)). This illustrates the importance of fusing readings from multiple sensors for associating proximity events with items (\secref{sec:behavior}). The biggest loss of accuracy comes from using only the vision sensors to detect items. RFID sensors perform the best, since RFID can accurately determine which item was selected\sepfootnote{rfid}. Even so, an RFID-only deployment has 12\% lower recall and 16\% lower precision. Of the sensor combinations, using weight and RFID sensors together comes closest to the recall performance of the complete system, losing only about 3\% in recall, but 10\% in precision.


\parab{Adversarial actions} \figref{fig:action_accuracy}(b) shows precision and recall for only those actions in which users tried to switch items. In these cases, \sys is able to achieve nearly 90\% precision and recall, while the best single sensor (RFID) has 7\% lower recall and 13\% lower precision, and the best 2-sensor combination (weight and RFID) has 5\% lower precision and recall. As expected, using a vision sensor or weight sensor alone has unacceptable performance because the vision sensor cannot distinguish between items that look alike and the weight sensor cannot distinguish items of similar weight.

\figref{fig:action_accuracy}(c) shows precision and recall for only those actions in which users tried to hide the hand from the camera when picking up items. In these cases, \sys estimates proximity events from the proximity of the ankle joint to the shelf (\secref{sec:behavior}) and achieves a precision of 80\% and a recall of 85\%. In the future, we hope to explore cross-camera fusion to be more robust to these kinds of events. Of the single sensors, weight and RFID both have more than 24\% lower recall and precision than \sys. Even the best double sensor combination has 12\% lower recall and 20\% lower precision.

Finally, \figref{fig:action_accuracy}(d) shows precision and recall only for those items in which the user trying to tamper with the weight sensors. In these cases, \sys is able to achieve nearly 87\% recall and 80\% precision. RFID, the best single sensor, has more than 10\% lower precision and recall, while predictably, vision and RFID have the best double sensor performance with 5\% lower recall and comparable precision to \sys.

In summary, \sys has slightly lower precision and recall for the adversarial cases and these can be improved with algorithmic improvements, its overall precision and recall on a trace with nearly 40\% adversarial actions is over 91\%. When we analyze only the non-adversarial actions, \sys has \textit{a precision of 95.8\% and a recall of 97.2\%}.


\parab{Taxonomy of \sys failures} 
\sys is unable to recall 19 of the 307 events in our trace. These failures fall into two categories: those caused by identity tracking, and those by action recognition. Five of the 19 failures are caused either by wrong face identification (2 in number), false pose detection (2 in number) (\figref{fig:eva_figs}(c)), or errors in pose tracking (one). The remaining failures are all caused by inaccuracy in action recognition, and fall into three categories. First, \sys uses color histograms to detect items (\secref{sec:behavior}), but these can be sensitive to lighting conditions (\eg a shopper takes an item from one shelf and puts it in another when the lighting condition is slightly different) and occlusion (\eg a shopper deposits an item into a group of other items which partially occlude the items). Incomplete background subtraction can also reduce the accuracy of item detection. Second, our weight scales were robust to noise but sometimes still could not distinguish between items of similar, but not identical, weight. Third, our RFID-to-proximity event association failed at times when the tag's RFID signal disappeared for a short time from the reader, possibly because the tag was temporarily occluded by other items. Each of these failure types indicates directions or future work for \sys.

\parab{Contextualizing the results} 
From the precision/recall results, it is difficult to know if \sys is within the realm of feasibility for use in today's retail stores. \sys's failures fall into two categories: \sys associates the wrong item with a shopper, or it associates an item with the wrong shopper. The first can result in inventory loss, the second in overcharging a customer. A survey of retailers~\cite{nrf} estimates the \textit{inventory loss ratio} (if a store's total sales are \$100, but \$110 worth of goods were taken from the store, the inventory loss rate is 10\%) in today's stores to be 1.44\%. In our experiments, \sys's failures result in only 0.79\% inventory loss. Another study~\cite{shopper_cost_rate} suggests that faulty scanners can result in up to 3\% overcharges on average, per customer. In our experiments, we see a 2.8\% overcharge rate. These results are encouraging and suggest that \sys may be with the realm of feasibility, but larger scale experiments are needed to confirm this. Additional investments in sensors and cameras, and algorithm improvements, could further improve \sys's accuracy.

\subsection{The Importance of Efficiency}
\label{sec:efficiency_eval}

\begin{figure}
\centering\includegraphics[width=0.85\columnwidth]{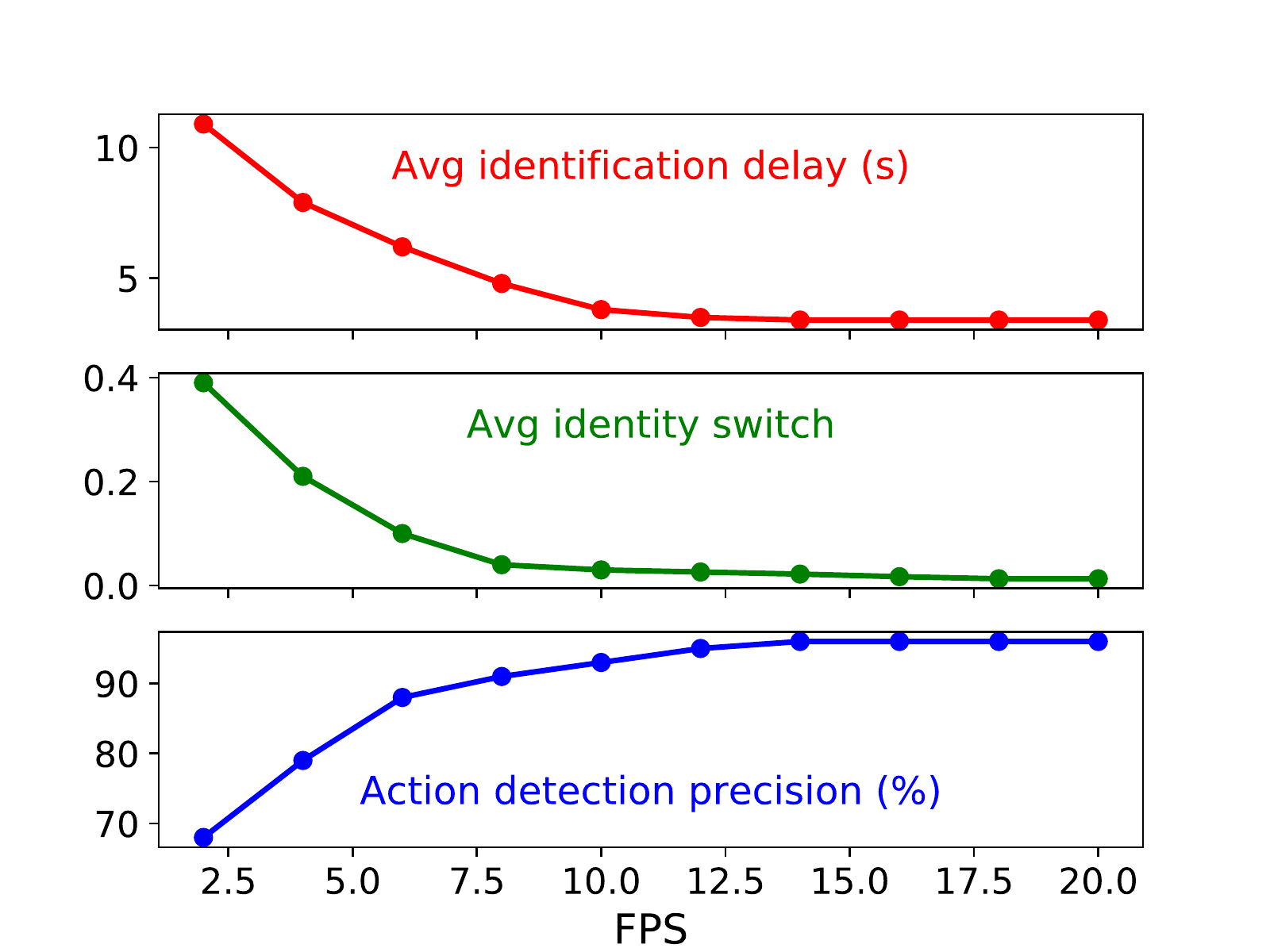}
\caption{\emph{\sys needs a frame rate of at least 10 fps for sufficient accuracy, reducing identity switches and identification delay.}}
\label{fig:fps_relation}
\end{figure}

\sys is designed to process data in near real-time so that customers can be billed automatically as soon as they leave the store. For this, computational efficiency is important to lower cost (\secref{sec:scale_eval}), but also to achieve high processing rates in order to maintain accuracy. 

\sepfootnotecontent{precision}{In this and subsequent sections, we focus on precision, since it is lower than recall (\secref{sec:accuracy_eval}), and so provides a better bound on \sys performance.}

\parab{Impact of lower frame rates}
If \sys is unable to achieve a high enough frame rate for processing video frames, it can have significantly lower accuracy. At lower frame rates, \sys can fail in three ways. First, a customer's face may not be visible at the beginning of the track in one camera. It usually takes several seconds before the camera can capture and identify the face. At lower frame rates, \sys may not capture frames where the shopper's face is visible to the camera, so it might take longer for it to identify the shopper. \figref{fig:fps_relation}(a) shows that this identification delay decreases with increasing frame rate approaching sub-second times at about 10 fps. Second, at lower frame rates, the shopper moves a greater distance between frames, increasing the likelihood of \textit{identity switches} when the tracking algorithm switches the identity of the shopper from one registered user to another. \figref{fig:fps_relation}(b) shows that the ratio of identity switches approaches negligible values only after about 8~fps. Finally, at lower frame rates, \sys may not be able to capture the complete movement of the hand towards the shelf, resulting in incorrect determination of proximity events and therefore reduced overall accuracy. \figref{fig:fps_relation}(c) shows precision\sepfootnote{precision} approaches 90\% only above 10~fps.


\parab{Infeasibility of a DNN-only architecture}
In \secref{sec:design} we argued that, for efficiency, \sys could not use separate DNNs for different tasks such as identification, tracking, and action recognition. To validate this argument, we ran the state-of-the-art open-source DNNs for each of these tasks on our data set. These DNNs were at the top of the leader-boards for various recent vision challenge competitions~\cite{coco_challenge, mot_challenge, mpii}. We computed both the average frame rate and the precision achieved by these DNNs on our data (\tblref{tbl:other_options}).

For face detection, our accuracy measures the precision of face identification. The OpenFace~\cite{amos2016openface} DNN can process 15 fps and achieve the precision of 95\%. For people detection, our accuracy measures the recall of bounding boxes between different frames. Yolo~\cite{yolo} can process at a high frame rate but achieves only 91\% precision, while Mask-RCNN~\cite{he2017mask} achieves 97\% precision, but at an unacceptable 5 fps. The DNNs for people tracking showed much worse behavior than \sys, which can achieve an identity switch rate of about 0.027 at 10~fps, while the best existing system, DeepSORT~\cite{wojke2017simpl} has a higher frame rate but a much higher identity switch rate. The fastest gesture recognition DNN is OpenPose~\cite{cao2017realtime} (whose body frame capabilities we use), but its performance is unacceptable, with low (77\%) accuracy. The best gesture tracking DNN, PoseTrack~\cite{iqbal2016PoseTrack}, has a very low frame rate.

Thus, today's DNN technology either has very low frame rates or low accuracy for individual tasks. Of course, DNNs might improve over time along both of these dimensions. However, even if, for each of the four tasks, DNNs can achieve, say, 20~fps and 95\% accuracy, when we run these on a single GPU, we can at best achieve 5~fps, and an accuracy of $0.95^4~=~0.81$. By contrast, \sys is able to process a single camera on a single GPU at over 15 fps (\figref{fig:scale_eval}), achieving over 90\% precision and recall (\figref{fig:action_accuracy}(a)).


\begin{table}[htbp]
\small
\centering
\begin{tabular}{|l|l|l|}
\hline
\rowcolor[HTML]{CBCEFB} 
\textit{\textbf{Face Detection}}              & \textbf{FPS} & \textbf{Accuracy}      \\ \hline
\textit{OpenFace~\cite{amos2016openface}}     & 15           & 95.1                   \\ \hline
\textit{RPN~\cite{hao2017scale}}              & 5.8          & 95.1                   \\ \hline
\rowcolor[HTML]{ECF4FF} 
\textit{\textbf{People detection}}          & \textbf{FPS} & \textbf{Accuracy}      \\ \hline
\textit{YOLO-9000~\cite{yolo}}              & 35           & 91.0                   \\ \hline
\textit{Mask-RCNN~\cite{he2017mask}}        & 5            & 97.4                   \\ \hline
\rowcolor[HTML]{D6D6D6} 
\textit{\textbf{People tracking}}           & \textbf{FPS} & \textbf{Avg ID switch} \\ \hline
\textit{MDP~\cite{HenschelLCR17}}           & 1.43         & 1.3                    \\ \hline
\textit{DeepSORT~\cite{wojke2017simpl}}     & 17           & 0.8                    \\ \hline
\rowcolor[HTML]{FFFFC7} 
\textit{\textbf{Gesture Recognition}}               & \textbf{FPS} & \textbf{Accuracy*}     \\ \hline
\textit{OpenPose~\cite{cao2017realtime}}            & 15.5         & 77.3                   \\ \hline
\textit{DeeperCut~\cite{insafutdinov2016deepercut}} & 0.09         & 88                     \\ \hline
\rowcolor[HTML]{FFCCC9} 
\textit{\textbf{Gesture Tracking}}              & \textbf{FPS} & \textbf{Avg ID switch} \\ \hline
\textit{PoseTrack~\cite{iqbal2016PoseTrack}}    & 1.6         & 1.8                    \\ \hline
\end{tabular}
\caption{\emph{State-of-the-art DNNs for many of \sys's tasks either have low frame rates or insufficient accuracy. (* Average pose precision on MPII Single Person Dataset)}}
\label{tbl:other_options}
\end{table}


\subsection{GPU multiplexing} 
\label{sec:scale_eval}

In the results presented so far, \sys processes each camera on a separate GPU. The bottleneck in \sys is pose detection, which requires about 63 ms per frame: our other components require less than 7 ms each (\tblref{tbl:time_breakdown}).

\begin{figure}[htbp]
\centering
\includegraphics[width=0.85\columnwidth]{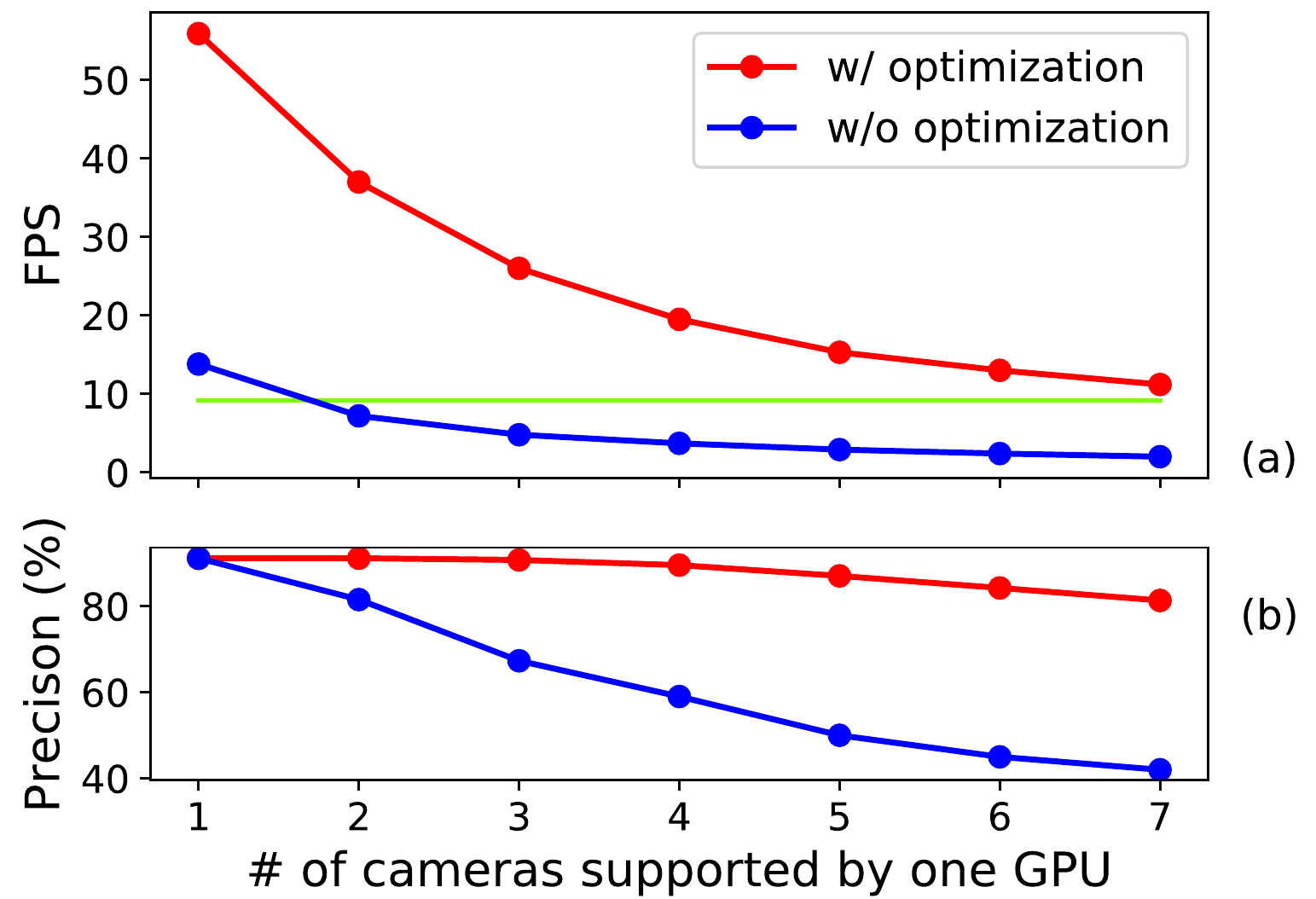}
\caption{\emph{GPU multiplexing can support up to 4 multiple cameras at frame rates of 10 fps or more, without noticeable lack of accuracy in action detection.}}
\label{fig:scale_eval}
\end{figure}

In \secref{sec:scale}, we discussed an optimization that uses a fast feature tracker to multiplex multiple cameras on a single GPU. This technique can sacrifice some accuracy, and we are interested in determining the sweet spot between multiplexing and accuracy. \figref{fig:scale_eval} quantifies the performance of our GPU multiplexing optimization. \figref{fig:scale_eval}(a) shows that \sys can support up to 4 cameras with a frame rate of 10 fps or higher with fast feature tracking; without it, only a single camera can be supported on the GPU (the horizontal line in the figure represents 10 fps). Up to 4 cameras, \figref{fig:scale_eval}(b) shows that the precision can be maintained at nearly 90\% (\ie negligible loss of precision). Without fast feature tracking, multiplexing multiple cameras on a single GPU reduces the effective frame rate at which each camera can be processed,  reducing accuracy for 4 cameras to under 60\%. Thus, with GPU multiplexing using fast feature tracking, \sys can reduce the investment in GPUs by 4$\times$.


\subsection{Evaluating Design Choices} 
\label{sec:choices}

In this section, we experimentally validate design choices and optimizations in identification and tracking.

\begin{table}[t]
\small
\centering
\begin{tabular}{|l|l|}
\hline
\textbf{Module}              & \textbf{Avg time per frame (ms)} \\ \hline
\textit{Pose detection}      & 63.3                             \\ \hline
\textit{Face detection}      & 4.1                              \\ \hline
\textit{Face identification} & 7                              \\ \hline
\textit{Pose tracking}       & 5.0                              \\ \hline
\end{tabular}
\caption{\emph{Pose detection is the bottleneck in \sys.}}
\label{tbl:time_breakdown}
\end{table}

\sepfootnotecontent{retrain}{Fast re-training is essential to minimize the time the customer needs to wait between registration and shopping.}

\parab{Identification}
Customer identification in \sys consists of three steps (\secref{sec:identity}): face detection, feature extraction, and feature classification. For face detection, \sys adjusts the bounding box from OpenPose output. It could have used the default OpenPose output box or run a separate neural network for face detection. \tblref{face_detector_opt} shows that our design choice preserves detection accuracy while being an order of magnitude faster. For feature extraction, we compared our ResNet face features with another face feature (FaceNet), with a neural net generated body feature, and with a body color histogram. \tblref{face_feature_opt} shows that our approach has the highest accuracy. Finally, for feature classification, we tried three approaches: comparing features' cosine distance, using kNN, or using a simple neural network. Their re-training time\sepfootnote{retrain}, running speed, and accuracy, are shown in \tblref{face_classifier_opt}. We can see that kNN has the best accuracy with retraining overhead of 2 s and classification overhead less than 2 ms.

\begin{table}[htbp]
\small
\centering
\begin{tabular}{|l|l|l|}
\hline
  & \textbf{Speed (ms/img)} & \textbf{Accuracy (\%)} \\ \hline
\textit{Adjusted box (Grab)} & 4.1                       & 95.1                                  \\ \hline
\textit{Original box from pose}        & \textless 1               & 83.0                                  \\ \hline
\textit{Box from DNN model}       & 93                        & 95.1                                                          \\ \hline
\end{tabular}
\caption{\emph{Adjusting the face bounding box in \sys has comparable accuracy to a neural network based approach, while having significantly lower overhead.}}
\label{face_detector_opt}
\end{table}

\vspace{-0.3in}

\begin{table}[htbp]
\small
\centering
\begin{tabular}{|l|l|}
\hline
\textbf{Method} & \textbf{Accuracy} \\ \hline
\textit{\sys's model (ResNet)}             & 95.1\%                            \\ \hline
\textit{FaceNet}                      & 89.4\%                            \\ \hline
\textit{Body deep feature}            & 51.2\%                            \\ \hline
\textit{Color histogram}              & 31.7\%                            \\ \hline
\end{tabular}
\caption{\emph{\sys's ResNet based face features have the highest accuracy.}}
\label{face_feature_opt}
\end{table}

\vspace{-0.2in}

\begin{table}[t]
\small
\centering
\begin{tabular}{|l|l|l|l|}
\hline
                                            & \textbf{Cosine Dist} & \textbf{kNN} & \textbf{Neural Net} \\ \hline
\textit{Retraining latency (s)}  & 0                    & 2.1          & 68.6                \\ \hline
\textit{Classification latency (ms)}       & 0.1*                 & 1.9          & 10.7                \\ \hline
\textit{Accuracy (\%)} & 75.6                 & 95.1         & 92.8  
\\ \hline
\end{tabular}
\caption{\emph{\sys's kNN-based algorithm has highest accuracy while having low retraining and classification latency. (* Cosine distance runtime is 0.1 ms per person)}}
\label{face_classifier_opt}
\end{table}

\vspace{0.1in}

\parab{Tracking}
Replacing our pose tracker with a bounding box tracker~\cite{wojke2017simpl} can result in below 50\% precision. Removing the limb association optimization drops precision by about 11\%, and removing the optimization that estimates proximity when the hand is not visible reduces precision by over 7\%. Finally, removing lazy tracking, which permits accurate tracking even in the presence of occlusions can reduce precision by over 15\%. Thus, each  optimization is necessary to achieve high precision.


\section{Related Work}
\label{sec:related}

We are not aware of published work on end-to-end design and evaluation of cashier-free shopping. 

\parab{Commercial cashier-free shopping systems}
Amazon Go was the first set of stores to permit cashier-free shopping. Several other companies have deployed demo stores, including Standard Cognition~\cite{std_cog}, Taobao~\cite{taobao}, and Bingobox~\cite{bingobox}. Amazon Go and Standard Cognition use deep learning and computer vision to determine shopper-to-item association (\cite{amzn_no_rfid, go-geek, go-nyt, std_cog_forbes}). Amazon Go does not use RFID \cite{go-nyt, go-geek} but needs many ceiling-mounted cameras. Imagr\cite{imagr} uses a camera-equipped cart to recognize the items put into the cart by the user. Alibaba and Bingobox use RFID reader to scan all items held by the customer at a "checkout gate" (\cite{ali_tech, bingo_tech}). \sys incorporates many of these elements in its design, but uses a judicious combination of complementary sensors (vision, RFID, weight scales).


\parae{Person identification} 
Person (re)-identification has used face features and body features.
Body-feature-based re-identification~\cite{bai2017scalable, chen2017beyond, zhao2017spindle} can achieve the precision of up to 80\%, insufficient for cashier-free shopping. Proprietary face feature based re-identification~\cite{face_id, yitu, sighthound_face, amazon_face_recog} can reach 99\% precision. Recent academic research using face features has achieved an accuracy of more than 95\% on public datasets, but such systems are either unavailable~\cite{hayat2017joint, chen2018face, zhao2018towards} or too slow~\cite{tran2017disentangled, masi2016pose}. \sys uses fast feature-based face re-identification with comparable accuracy while using a pose tracker to accurately bound the face (\secref{sec:identity}).
 
\parae{People tracking} Bounding box based trackers~\cite{ristani2018features, wojke2017simpl, liu2018tar} can track shopper movement, but can be less effective in crowds (\secref{sec:efficiency_eval}) since they do not detect limbs and hands. Some pose trackers~\cite{Iqbal_CVPR2017, insafutdinov2017} can do pose detection and tracking at same time, but are too slow for \sys (\secref{sec:efficiency_eval}) which uses a skeleton-based pose tracker both for identity tracking and gesture recognition.

\parae{Action detection} Action detection is an alternative approach to identifying shopping actions. Publicly available state-of-the-art DNN-based solutions \cite{kalogeiton17iccv, sun2018optical, heilbron2017scc, dave2017predictive} have not yet been trained for shopping actions, so their precision and recall in our setting is low. 


\parae{Item detection and tracking} Prior work has explored item identification using Google Glass~\cite{ha2014towards} but such devices are not widely deployed. RFID tag localization can be used for item tracking~\cite{shangguan2015relative, shangguan2017, jiang2018orientation} but that line of work does not consider frequent tag movements, tag occlusion, or other adversarial actions. Vision-based object detectors~\cite{he2017mask, chen2018domain, redmon2018yolov3} can be used to detect items, but need to be trained for shopping items and can be ineffective under occlusions and poor lighting (\secref{sec:accuracy_eval}). Single-instance object detection scales better for training items but has low accuracy~\cite{karlinsky2017fine,held2016robust}.


\section{Conclusion}
\label{sec:concl}

Cashier-free shopping systems can help improve the shopping experience, but pose significant design challenges. \sys is a cashier-free shopping system that uses a skeleton-based pose tracking DNN as a building block, but develops lightweight vision processing algorithms for shopper identification and tracking, and uses a probabilistic matching technique for associating shoppers with items they purchase. \sys achieves over 90\% precision and recall in a data set with up to 40\% adversarial actions, and its efficiency optimizations can reduce investment in computing infrastructure by up to 4$\times$. Much future work remains including obtaining results from longer-term deployments, improvements in robust sensing in the face of adversarial behavior, and exploration of cross-camera fusion to improve \sys's accuracy even further.



{\small
\bibliographystyle{abbrv}  
\bibliography{references}
}



\end{document}